\documentclass[lettersize,journal]{IEEEtran}
\usepackage{amsmath,amsfonts}
\usepackage{algorithmic}
\usepackage{algorithm}
\usepackage{array}
\usepackage[caption=false,font=normalsize,labelfont=sf,textfont=sf]{subfig}
\usepackage{textcomp}
\usepackage{stfloats}
\usepackage{url}
\usepackage{verbatim}
\usepackage{graphicx}
\usepackage{multirow}
\usepackage{makecell}
\usepackage{cite}
\usepackage{textcomp} 
\DeclareUnicodeCharacter{2082}{\ensuremath{_2}}

\begin{document}
\bstctlcite{IEEEexample:BSTcontrol}

\title{HiGFRL: Hierarchical Graph Fusion-Driven Reinforcement Learning for Dependency-Aware Task Scheduling in Heterogeneous Cloud}

\author{Tiangang Li, Shi Ying, Xiangbo Tian
\thanks{Tiangang Li, Shi Ying and Xiangbo Tian are with the School of Computer Science, Wuhan University, Wuhan 430072, China. (Email: tiangangli@whu.edu.cn; yingshi@whu.edu.cn; tianxiangbo@whu.edu.cn)}
\thanks{This work has been submitted to the IEEE for possible publication. Copyright may be transferred without notice, after which this version may no longer be accessible.}
\thanks{(Corresponding author: Shi Ying.)}
}



\maketitle

\begin{abstract}
Online scheduling of dependency-aware tasks in heterogeneous cloud clusters is a fundamental yet challenging problem due to the complex interplay between DAG topologies and multi-dimensional resource constraints. While Deep Reinforcement Learning (DRL) has shown promise, existing Graph Neural Network-based approaches often struggle to efficiently model high-order topological dependencies and suffer from loose coupling between task and resource states, leading to myopic scheduling decisions. To address these limitations, we propose HiGFRL, a Hierarchical Graph Fusion-Driven Reinforcement Learning framework. HiGFRL constructs a novel three-level state representation comprising a Static Hypergraph, a Dynamic Global Graph, and a Local Bipartite Graph to explicitly model the interplay between task dependencies and real-time cluster dynamics. Specifically, we design a fusion-driven dual-network architecture to optimize reinforcement learning decision-making, where a Context Fusion Allocator integrates local bipartite matching features with fused global context to execute precise task-to-node allocation, and a Global State Evaluator leverages the global dynamic graph representation to accurately estimate expected long-term cumulative reward. Furthermore, we incorporate a topology-prior-guided hybrid reward mechanism that distills static topological priors into the learning process to accelerate convergence. Extensive experiments using real-world Alibaba cluster traces demonstrate that HiGFRL significantly outperforms heuristics and DRL baselines. Specifically, in challenging large-scale high-load scenarios, HiGFRL reduces the Makespan by up to 32.55\%, and optimizes the average task flow time and average task wait time by 13.58\% and 13.79\%, respectively. The experimental results confirm that HiGFRL not only significantly improves cluster job throughput but also ensures superior Quality of Service by substantially reducing queuing delays.

\end{abstract}


\section{Introduction}
\IEEEPARstart{C}{loud} computing infrastructure is undergoing a rapid evolution from homogeneous clusters to highly heterogeneous environments \cite{ref31},\cite{ref32}. These environments integrate diverse resource configurations, such as different specifications of CPU and memory combinations, and non-uniform processing capabilities, thoroughly revolutionizing the execution paradigm of cloud service applications \cite{ref25},\cite{ref30}. Concurrently, the complexity of cloud workloads has increased significantly \cite{ref26},\cite{ref28}. To unify the representation of diverse batch workloads in production environments, recent studies and large-scale traces, such as Alibaba cluster data, have widely adopted a standardized Job-Task-Instance hierarchical abstraction \cite{ref1}, \cite{ref27},\cite{ref29}. In this paradigm, a Job represents a complete workflow modeled as a Directed Acyclic Graph (DAG), a Task serves as a logical execution unit within the DAG, and an Instance is the physical execution entity carrying the actual computational load.

This hierarchical paradigm introduces a critical challenge: Instance-level Online Dependency-aware Scheduling. Unlike traditional DAG scheduling that treats tasks as atomic units, the scheduler must make real-time decisions to map dynamically arriving streams of fine-grained instances onto massive heterogeneous resources, while strictly satisfying precedence constraints derived from task dependencies, multi-dimensional resource capacity constraints, and assignment uniqueness constraints. The objective is to jointly optimize multiple conflicting performance metrics, specifically minimizing the average task flow time, task wait time, and overall makespan. This is a typical NP-hard problem that requires the scheduling algorithm to strike a delicate balance between long-term topological planning and immediate response for resource allocation in highly dynamic and uncertain environments \cite{ref33},\cite{ref34},\cite{ref35}.

Traditional solutions primarily rely on heuristic algorithms \cite{ref2}, \cite{ref3}, \cite{ref4}, \cite{ref5}, \cite{ref6}, \cite{ref7}. Static list scheduling approaches prioritize tasks based on topological ranks. Although computationally efficient, they lack the adaptability to cope with the stochastic arrival patterns of runtime workloads. Dynamic heuristics, such as Tetris, adopt greedy strategies for multi-dimensional resource packing. However, they are often limited by myopia, tending to block future critical path tasks in pursuit of short-term resource utilization. To overcome these rigidities, Deep Reinforcement Learning (DRL) has emerged as a promising alternative. Nevertheless, early DRL-based approaches utilizing MLPs or CNNs, such as DeepRM, struggled to effectively capture the non-Euclidean topological dependencies inherent in DAGs.

Recognizing this, recent research has shifted towards Graph Neural Network (GNN)-based DRL paradigms \cite{ref8}, \cite{ref9}, \cite{ref10}, \cite{ref11}, \cite{ref36}. Pioneering works like Decima and READYS have utilized Graph Convolutional Networks (GCNs) to encode DAG structures and combined them with DRL for decision-making, becoming the mainstream direction for solving complex scheduling problems \cite{ref1-12,ref2-13}. More advanced approaches have introduced Graph Attention Networks (GATs) to capture pairwise task correlations \cite{ref2-11,ref2-12}. Despite the progress, existing GNN-based approaches still suffer from two critical defects when applied to complex heterogeneous environments.

Firstly, conventional directed graph structures have intrinsic defects in capturing many-to-one synchronization dependencies. Real-world cloud workflows widely contain Join patterns, where the readiness of a task is strictly constrained by the simultaneous completion of multiple predecessor tasks. Existing approaches typically model these dependencies as pairwise edges, forcing GNNs to rely on multi-hop message passing to aggregate upstream information. This iterative process often leads to severe over-smoothing and signal dilution, making it difficult for the scheduler to precisely perceive the dependency aggregation constraints of tasks deep in the dependency chain. In contrast, introducing the Hypergraph is crucial because hyperedges can directly and losslessly aggregate features from all predecessors in a single convolution operation, thereby preserving high-order synchronization semantics and providing the scheduler with a clearer topological vision.

Secondly, existing state representation and fusion mechanisms suffer from a severe mismatch in aligning topological importance with resource availability. Existing approaches, such as MODRL, typically encode topological features and resource features independently and rely on simple concatenation to combine them \cite{ref2-12}. Such approaches lack a decoupling and fusion strategy to explicitly model the high-order interactions between the topological importance of a task within the static view and the heterogeneous capabilities of specific nodes within the dynamic view. For example, the model inherently treats resource state merely as an additional node attribute and fails to simultaneously perceive the intrinsic connection between a task being on the critical path and a node possessing high available resource capacity and strong processing capability. This leads to decision myopia: the scheduler might maximize current resource packing efficiency but fail to reserve high-performance resources for future critical path tasks, ultimately damaging long-term performance.

To address these challenges, we propose HiGFRL, a hierarchical graph fusion-driven reinforcement learning framework. HiGFRL aims to synergistically optimize fine-grained topology-aware and resource-aware decisions through a multi-view representation learning paradigm. 

\noindent \textbf{Contributions.} The main contributions of this study are summarized as follows:

\begin{itemize}

\item{{\bf Hierarchical Multi-View State Construction:}  We propose a novel multi-level state representation mechanism, comprising the static task dependency hypergraph for efficiently capturing high-order dependencies and synchronization constraints, the dynamic global cluster graph for real-time perception of cluster-wide resource posture, and the local bipartite matching graph for optimizing microscopic matching decisions. This hierarchical structure achieves explicit decoupling and semantic enhancement of the system state.}

\item{{\bf Fusion-Driven Dual-Stream Architecture:}  A fusion-driven dual-network architecture is designed. The context-aware matching actor network employs a cross-level fusion mechanism to integrate local bipartite matching features with the broadcasted global context, executing precise instance-to-node assignments. Meanwhile, the global state-value critic network utilizes the global dynamic graph representation to accurately estimate the expected long-term cumulative reward, effectively reducing the variance of policy gradients.}

\item{{\bf Structural-Semantic Decoupling and Fusion:} By hierarchically decoupling the logical task topology from physical resource states, HiGFRL effectively alleviates the semantic entanglement and over-smoothing issues inherent in flat graph representations. This mechanism successfully fuses topological importance with resource availability, ensuring precise identification of critical path bottlenecks by preventing their topological signals from being diluted by noisy resource fluctuations.}

\item{{\bf Extensive Evaluation:} HiGFRL is evaluated using large-scale real-world Alibaba cluster traces. Experimental results demonstrate that HiGFRL significantly outperforms baselines across various metrics. Specifically, in the large-scale high-load scenario, HiGFRL reduces the makespan by 32.55\%. In the extreme load scenario, it reduces the average task flow time by 13.58\% and the average task wait time by 13.79\%. These results indicate that HiGFRL exhibits superior performance in both enhancing job throughput and improving user experience.}

\end{itemize}

The rest of this paper is organized as follows. Section 2 introduces related work. Section 3 provides a detailed definition of the research problem and the system model. Section 4 presents the design of the HiGFRL framework. Section 5 presents the experimental design and results analysis. Section 6 concludes the research findings and future work.

\section{RELATED WORK}

In modern cloud and distributed systems, complex applications typically manifest as DAG workflows with strict task dependencies. Dependency-aware scheduling aims to optimize system metrics by rationally allocating computing resources across DAG nodes. To address this problem, existing research can be broadly categorized into heuristic-based approaches and DRL-based approaches.

\noindent {\bf{Heuristic-Based Approaches.}} 
Regarding single-objective or constrained optimization, Wu \emph {et al}. \cite{ref2} propose the L-ACO and ProLiS algorithms, utilizing probabilistic upward rank for deadline-constrained subtask allocation to balance cost and time under strict constraints. For edge-cloud environments, Lin \emph {et al}. \cite{ref3} introduce a cost-driven strategy that reduces data transmission costs by merging cut edges and scheduling task sequences on partial critical paths. Zhang \emph {et al}. \cite{ref4} propose the BCWS algorithm to minimize overall completion time across heterogeneous VM and serverless resources under budget constraints.
To address conflicting practical objectives, multi-objective optimization is widely studied. Ismayilov \emph {et al}. \cite{ref5} propose NN-DNSGA-II, a neural network-assisted evolutionary algorithm simultaneously optimizing six dimensions, including completion time, execution cost, energy consumption, load balancing, system reliability, and resource utilization. Qin \emph {et al}. \cite{ref6} develop RA-MOMA, enhancing scheduling robustness via critical path analysis and local resource optimization strategies. Additionally, Sun \emph {et al}. \cite{ref7} apply a multi-tree genetic programming approach to automatically evolve rules for task and resource selection in dynamic multi-cloud environments.
Overall, while heuristic approaches achieve satisfactory performance in specific stable scenarios by leveraging domain knowledge, they generally struggle to adapt to highly dynamic workloads and heterogeneous resources. They often require tedious manual parameter tuning, thus lacking generalizability.

\noindent {\bf{DRL-based Approaches.}} To overcome the poor adaptability of heuristics in dynamic environments, recent research actively utilizes DRL for adaptive dependency-aware scheduling, focusing on neural networks to process complex DAG topologies. Early DRL approaches rely on basic architectures like MLPs or CNNs. For instance, Dong \emph{et al}. \cite{ref8} formulate scheduling as a Markov Decision Process (MDP) and propose RLWS, an Actor-Critic approach with iterative local rescheduling. Cheng \emph{et al}. \cite{ref9} introduce H2O-Cloud, a hierarchical online framework enabling pre-training-free DRL scheduling, significantly easing practical deployment.

With the rise of GNNs, Graph Reinforcement Learning (GRL) approaches directly modeling non-Euclidean DAG structures have become a research hotspot \cite{ref37}. Decima \cite{ref1-12} pioneers this by embedding job graphs via GNNs and training with REINFORCE, significantly improving completion times compared to classical heuristics. Hu \emph{et al}. \cite{ref10} propose Spear, combining Monte Carlo Tree Search (MCTS) with DRL to optimize scheduling under complex dependencies and heterogeneous demands, yielding notable improvements. Improved GNN architectures further enhance DAG modeling accuracy. Drag-JDEC \cite{ref11} employs GATs for edge computing scheduling, achieving a remarkable performance boost. GA-DRL \cite{ref2-11} utilizes multi-head GATs with bidirectional aggregation and non-uniform sampling for strong generalization on unseen DAGs. Wang \emph{et al}. \cite{ref2-12} propose a spatio-temporal GNN with a Set Transformer for multi-objective scheduling. READYS \cite{ref2-13} extracts topological features via GCNs and constructs graph-level states via global pooling for adaptive dynamic scheduling. Moreover, the introduction of attention mechanisms significantly improves scheduling decision quality. Wang \emph{et al}. \cite{ref2-14} combine GATs with self-attention MLPs, creatively mapping continuous actions to discrete ones via k-d trees. SPN-CWS \cite{ref2-15} uses a self-attention policy network to capture global VM contexts, trained via evolutionary strategy-assisted RL. SpotDAG \cite{ref2-16} introduces self-attention with output masking to avoid invalid actions, ensuring deadlines while minimizing costs via spot instances.

Scenario-specific approaches continually enrich this field. Koslovski \emph{et al}. \cite{ref2-17} design an Actor-Critic scheduler to adaptively select heuristic policies based on real-time states. Co-ScheRRL \cite{ref2-18} targets co-located scenarios using self-attention and DRL relational reasoning. Xue \emph{et al}. \cite{ref2-19} reduce Yarn cluster job latency by utilizing resource queue idle windows. Dong \emph{et al}. \cite{ref2-20} optimize Storm streaming workloads via weighted GNN embeddings. Shu \emph{et al}. \cite{ref2-21} propose a multi-policy MCTS approach to adaptively adjust search strategies.

In summary, while existing GRL-based scheduling approaches demonstrate immense advantages in task dependency modeling and structural feature extraction, an in-depth analysis reveals several limitations. First, most approaches assume static or single-workflow graphs, struggling to cope with complex real-world environments characterized by stochastic task arrivals and concurrent workflow execution. Furthermore, Actor and Critic networks in these frameworks frequently share the same representation layer, restricting their capability to differentially model local task features and global system states. A summary of the main related works on DRL-based approaches is presented in Table \ref{tab:table_com}. 


\begin{table*}[!t]
\caption{Comparison of Main Related DRL-Based Dependency-Aware Task Scheduling Approaches.\label{tab:table_com}}
\centering
\begin{tabular}{p{1.0in}<{\raggedright}p{1.0in}<{\raggedright}p{1.15in}<{\raggedright}p{1.35in}<{\raggedright}p{1.5in}<{\raggedright}}
\hline
References & Topology Modeling & Representation Architecture & Resource-Task Fusion & RL Algorithm\\
\hline
\\
Grinsztajn \emph{et al}. \cite{ref2-13} & Flat GCN & Coupled & Global Pooling \& Concat & Advantage Actor-Critic \\
Yu \emph{et al}. \cite{ref11} & Flat GAT & Coupled & Simple Attribute Concat & DQN \\
Liu \emph{et al}. \cite{ref2-11} & Bi-directional GAT & Coupled & Simple Attribute Concat & Double DQN \\
Wang \emph{et al}. \cite{ref2-12} & Spatio-Temporal GNN & Coupled & Set Transformer & Dueling Double DQN \\
Lin \emph{et al}. \cite{ref2-16} & Self-Attention & Coupled & Simple Attribute Concat & Proximal Policy Optimization \\
\bf{Our work} & \bf{High-Order Hypergraph Convolution} & \bf{Decoupled Hierarchical Multi-View} & \bf{Cross-Level Feature Fusion} & \bf{Hierarchical Graph Fusion-Driven Dual-Stream Actor-Critic}\\
\hline
\end{tabular}
\end{table*}

\section{System Model and Problem Definition}

\subsection{System Model}

This section formally defines the scheduling environment, process, constraints, and optimization objectives. Specifically, we formulate the problem of scheduling multi-instance, dependency-aware tasks onto heterogeneous virtual machine clusters. For readability, Table \ref{tab:key_notations} summarizes the key notations used in the proposed system model.

\begin{table}[!t]
\caption{Summary of Key Notations in System Model}
\label{tab:key_notations}
\centering
\begin{tabular}{c l}
\hline
\textbf{Notation} & \textbf{Definition}\\
\hline
$\mathcal{M}$ & Set of heterogeneous VMs in the cluster\\
$M_k$ & The $k$-th VM in $\mathcal{M}$, where $k \in \{1, \ldots, |\mathcal{M}|\}$\\
$\mathcal{D}$ & Set of resource dimensions (e.g., $\{CPU, Mem\}$)\\
$\mathbf{C}_k$ & Total resource capacity vector of VM $M_k$\\
$\mathbf{A}_k(t)$ & Available resource vector of VM $M_k$ at time step $t$\\
$s_k$ & Processing speed coefficient of VM $M_k$\\
$\mathcal{J}$ & Set of streaming jobs arriving over time\\
$J_i$ & The $i$-th Job, modeled as a DAG $J_i=(\mathcal{T}_i, \mathcal{E}_i)$\\
$\mathcal{T}_i$ & Set of tasks within Job $J_i$\\
$\mathcal{E}_i$ & Set of precedence constraints (directed edges) in Job $J_i$\\
$T_{i,j}$ & The $j$-th Task of Job $J_i$\\
$\text{pred}(T_{i,j})$ & Set of direct predecessors of Task $T_{i,j}$\\
$a_{i,j}$ & Arrival time of Task $T_{i,j}$\\
$n_{i,j}$ & Number of instances contained in Task $T_{i,j}$\\
$d_{i,j}$ & Benchmark execution duration of a single instance of $T_{i,j}$\\
$\mathbf{r}_{i,j}$ & Resource demand vector of a single instance of $T_{i,j}$\\
$x_{i,j,l,k}(t)$ & Decision variable\\
$\delta_{i,j,k}$ & Actual execution duration of an instance on VM $M_k$\\
\hline
\end{tabular}
\end{table}

\noindent {\bf{(1) System Model.}} The system consists of a heterogeneous computing cluster and a dynamically arriving workload of jobs with precedence constraints.

\noindent {\bf{Compute Cluster.}} The cluster comprises a set of heterogeneous Virtual Machines (VMs), denoted as $\mathcal{M}=\{M_1, M_2, \ldots, M_{|\mathcal{M}|}\}$. The system considers a set of $|\mathcal{D}|$ distinct resource dimensions, indexed by $\mathcal{D}=\{1,2,\ldots,|\mathcal{D}|\}$. Each VM $M_k \in \mathcal{M}$ is characterized by a resource capacity vector $\mathbf{C}_k \in \mathbb{R}_+^{|\mathcal{D}|}$, where $C_k^d$ represents the total capacity of VM $M_k$ on resource dimension $d \in \mathcal{D}$. At time step $t$, the state of VM $M_k$ includes its available resource vector $\mathbf{A}_k(t) \in \mathbb{R}_+^{|\mathcal{D}|}$, where $A_k^d(t) \le C_k^d$ is the amount of available resources. Additionally, each VM $M_k$ possesses a processing speed coefficient $s_k \in \mathbb{R}^+$, which varies across different types of VMs.

\noindent {\bf{Workload Model.}} The workload consists of dependency-aware tasks arriving over time. These tasks are aggregated into a set of streaming jobs, $\mathcal{J}=\{J_1,J_2,\ldots,J_{|\mathcal{J}|}\}$, where each job forms a DAG encapsulating their precedence constraints.

{\bf{Job:}} Each job $J_i \in \mathcal{J}$ is modeled as a DAG, $J_i=(\mathcal{T}_i, \mathcal{E}_i)$. $\mathcal{T}_i=\{T_{i,1}, T_{i,2}, \ldots, T_{i,|\mathcal{T}_i|}\}$ is the set of tasks within job $J_i$. $\mathcal{E}_i \subseteq \mathcal{T}_i \times \mathcal{T}_i$ is the set of precedence constraints. A directed edge $(T_{i,j}, T_{i,k}) \in \mathcal{E}_i$ indicates that task $T_{i,j}$ must be completed before task $T_{i,k}$ can start. The set of direct predecessors of $T_{i,k}$ is denoted as $\text{pred}(T_{i,k})$.

{\bf{Task:}} Each task $T_{i,j} \in \mathcal{T}_i$ is defined by a tuple $T_{i,j}=(a_{i,j}, n_{i,j}, d_{i,j}, \mathbf{r}_{i,j})$, where $a_{i,j} \in \mathbb{R}_+$ is the arrival time of the task, $n_{i,j} \in \mathbb{Z}^+$ is the number of instances contained in the task, $d_{i,j} \in \mathbb{R}_+$ is the benchmark execution duration of a single instance, and $\mathbf{r}_{i,j} \in \mathbb{R}_+^{|\mathcal{D}|}$ is the resource demand vector of a single instance.

\noindent {\bf{(2) Scheduling Process and Constraints.}} The scheduling process is modeled as a sequence of decisions made at a discrete event timeline $(t_1, t_2, \ldots, t_{Z})$, where $Z$ represents the total number of decision steps. At any decision moment $t \in \{t_1, t_2, \ldots, t_{Z}\}$, the system needs to execute a scheduling action.

\noindent {\bf{Scheduling Decision Variables.}} The policy $\pi$ makes a decision at time $t$ by setting variables $x_{i,j,l,k}(t) \in \{0,1\}$, which indicate whether the $l$-th instance of task $T_{i,j}$ is assigned to VM $M_k$.

\begin{equation}
\label{eq1}
  x_{i,j,l,k}(t) = \begin{cases} 1, & \text{if instance } l \text{ of } T_{i,j} \text{ is assigned to } M_k \text{ at } t \\ 0, & \text{otherwise} \end{cases}
\end{equation}

\noindent where $l \in \{1, \ldots, n_{i,j}\}$ and $k \in \{1, \ldots, |\mathcal{M}|\}$.

\noindent {\bf{Execution Model.}} The actual execution duration of an instance on VM $M_k$, denoted as $\delta_{i,j,k}$, depends on the speed coefficient of the VM.

\begin{equation}
\label{eq2}
  \delta_{i,j,k} = \frac{d_{i,j}}{s_k}
\end{equation}

\noindent {\bf{Scheduling Feasibility Constraints.}} To ensure the validity of scheduling decisions, the process must strictly adhere to the following task dependency and resource capacity constraints:

{\bf{Precedence and Readiness Constraints.}} Task $T_{i,j}$ is eligible for scheduling at time $t$ only if it has arrived and all its predecessors have been completed. Let $FT_p$ be the finish time of task $T_p$. A decision for $T_{i,j}$ is valid only if the following condition is met.

\begin{equation}
\label{eq3}
    t \ge \max \left( \{a_{i,j}\} \cup \{ FT_p \mid T_p \in \text{pred}(T_{i,j}) \} \right)
\end{equation}

The earliest moment satisfying inequality Eq. \eqref{eq3} is defined as the ready time of the task, upon which the task enters the ready queue awaiting scheduling.

{\bf{Resource Constraints.}} A scheduling decision $x_{i,j,l,k}(t)=1$ is feasible only if the target VM has sufficient resources at the time of assignment. $\mathbf{r}_{i,j} \le \mathbf{A}_k(t)$ implies that for all dimensions $d \in \mathcal{D}$, $r_{i,j}^d \le A_k^d(t)$.

{\bf{Resource Dynamics.}} Upon a feasible assignment $x_{i,j,l,k}(t)=1$, the resources allocated on VM $M_k$ are immediately reserved for the entire actual duration of the instance.

\begin{equation}
\label{eq4}
    \mathbf{A}_k(\tau) \leftarrow \mathbf{A}_k(\tau) - \mathbf{r}_{i,j}, \quad \forall \tau \in [t, t + \delta_{i,j,k})
\end{equation}

{\bf{Instance Assignment Constraints.}} Each instance must be successfully scheduled exactly once.

\begin{equation}
\label{eq5}
    \sum_{t} \sum_{k=1}^{|\mathcal{M}|} x_{i,j,l,k}(t) = 1, \quad \forall i,j,l
\end{equation}

\begin{figure*}[!t]
\centering
\includegraphics[width=0.9\textwidth]{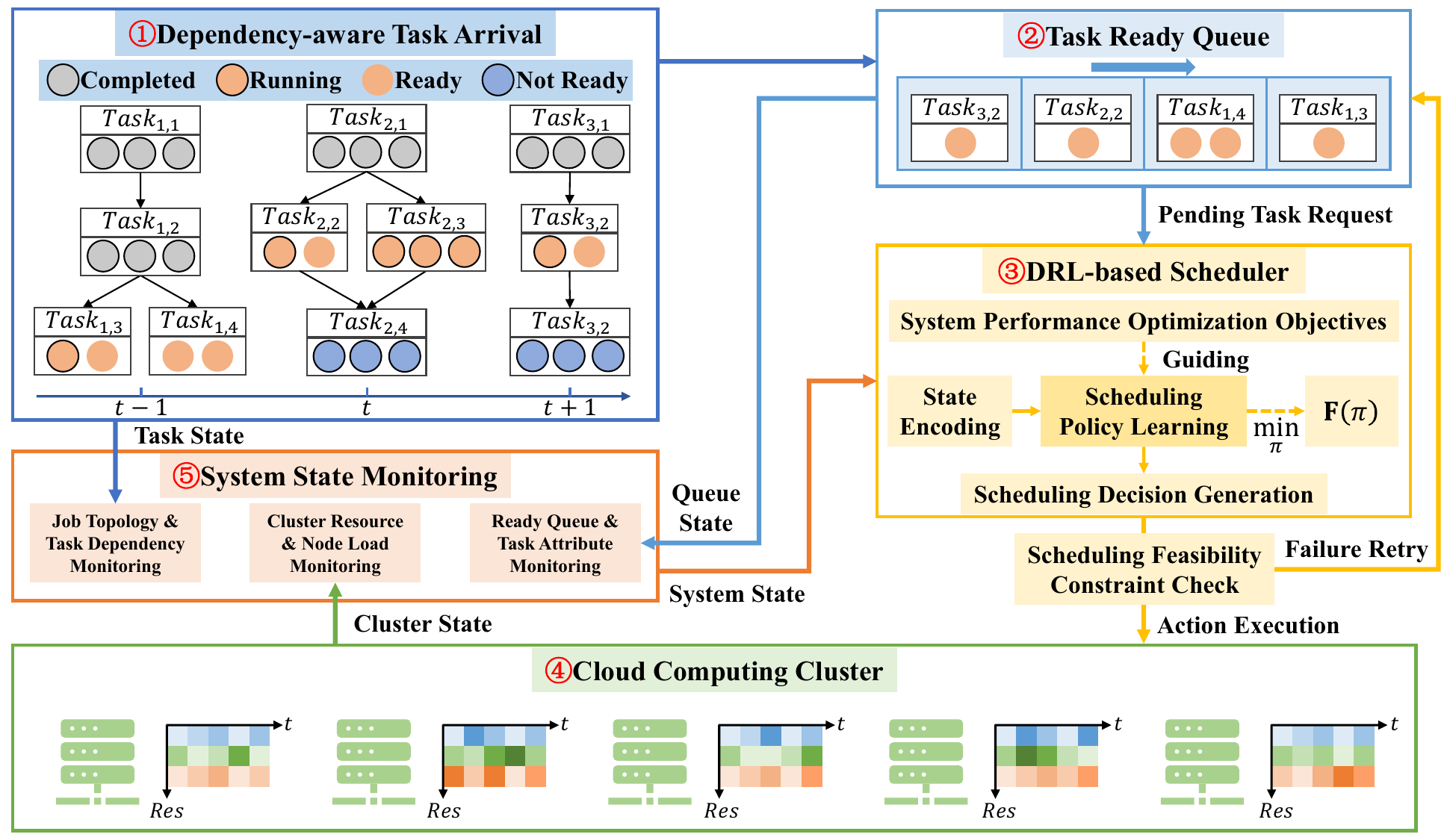}
\caption{System Model for Dependency-Aware Task Scheduling.}
\label{fig-sys}
\end{figure*}

The system model for dependency-aware task scheduling is illustrated in Fig. \ref{fig-sys}.

\noindent {\bf{(3) Problem Definition and Optimization Objectives.}}
The objective of the scheduling problem is to determine an optimal policy $\pi^*$ that generates a sequence of decisions to minimize a vector of performance objectives.
Let $ST_{i,j,l}$ be the start time and $FT_{i,j,l}$ be the completion time of the $l$-th instance of task $T_{i,j}$. Note that $FT_{i,j,l} = ST_{i,j,l} + \delta_{i,j,k}$.
The completion time of task $T_{i,j}$ is $FT_{i,j} = \max_l \{ FT_{i,j,l} \}$.
The start time of task $T_{i,j}$ is $ST_{i,j} = \min_l \{ ST_{i,j,l} \}$.
The task ready time, defined as the moment the task enters the ready queue, is denoted as $a_{i,j}^{ready}$. The optimization objective of the dependency-aware task scheduling problem is to find a policy $\pi$ to minimize the objective function $\mathbf{F}(\pi)$.

\begin{equation}
\label{eq6}
    \min_{\pi} \mathbf{F}(\pi) = (f_1(\pi), f_2(\pi), f_3(\pi))
\end{equation}

\noindent {\bf{Objective Functions:}}

{\bf{$f_1(\pi)$ (Average Task Flow Time):}} The average time spent by tasks in the system, from arrival to completion.

\begin{equation}
\label{eq7}
    f_1(\pi) = \frac{1}{\sum_{i} |\mathcal{T}_i|} \sum_{J_i \in \mathcal{J}} \sum_{T_{i,j} \in \mathcal{T}_i} (FT_{i,j} - a_{i,j})
\end{equation}

{\bf{$f_2(\pi)$ (Average Task Wait Time):}} The average time tasks wait in the ready queue for resources.

\begin{equation}
\label{eq8}
    f_2(\pi) = \frac{1}{\sum_{i} |\mathcal{T}_i|} \sum_{J_i \in \mathcal{J}} \sum_{T_{i,j} \in \mathcal{T}_i} (ST_{i,j} - a_{i,j}^{ready})
\end{equation}

{\bf{$f_3(\pi)$ (Makespan):}} The completion time of the entire workload.

\begin{equation}
\label{eq9}
    f_3(\pi) = \max_{J_i \in \mathcal{J}, T_{i,j} \in \mathcal{T}_i} \{ FT_{i,j} \}
\end{equation}

This problem is NP-hard. The objective is to find an optimal policy $\pi^*$ to minimize the objective function $\mathbf{F}(\pi)$ under precedence constraints, resource constraints, and assignment constraints.

\section{Approach}

We propose HiGFRL, a hierarchical graph fusion-driven reinforcement learning framework designed to address the complexity of dependency-aware scheduling in heterogeneous clouds. Unlike existing approaches relying on flat state representations or loosely coupled graph embeddings, HiGFRL introduces a multi-view representation learning paradigm. It explicitly decouples the system state into static topological structures and dynamic resource contexts, which are subsequently reintegrated through a fusion-driven dual-stream network architecture.

In this section, the dependency-aware scheduling problem is first modeled as a MDP. Then, the hierarchical multi-view state abstraction and the neural network architecture designed to capture high-order interactions between task dependencies and heterogeneous resources are detailed. Finally, the end-to-end joint optimization and training mechanism based on dual-stream decision heads is described.

\subsection{Graph Fusion-Driven Reinforcement Learning Model Construction}

The online scheduling problem of dependency-aware tasks in heterogeneous cloud environments is formalized as a sequential decision-making process, aiming to optimize long-term system performance objectives through a series of discrete scheduling actions. The process is modeled as a MDP defined by the tuple $(\mathcal{S}, \mathcal{A}, \mathcal{R}, \mathcal{P}, \gamma)$.
To handle the continuous-time dynamics of task arrivals and execution completions, this study adopts an event-driven mechanism to discretize the decision timeline. A decision step $t$ is defined as the moment the scheduler must make an assignment decision, typically triggered by the following event: a new task instance $T_{target}$ satisfies all predecessor dependencies and enters the ready queue, and there are available virtual machines in the cluster meeting its resource requirements. In this framework, the reinforcement learning agent (scheduler) observes the current system state $s_t \in \mathcal{S}$ at each decision step $t$, selects an action $a_t$ from the action space $\mathcal{A}$ according to the policy $\pi(a_t|s_t)$ to assign the task instance to a specific virtual machine. Subsequently, the environment transitions to the next state $s_{t+1}$ based on system dynamics and returns an immediate reward $r_t \in \mathcal{R}$ to the agent. Through this interaction process, the agent aims to learn an optimal policy $\pi^*$ to maximize the cumulative discounted return. 

\begin{figure}[!t]
\centering
\includegraphics[width=0.45\textwidth]{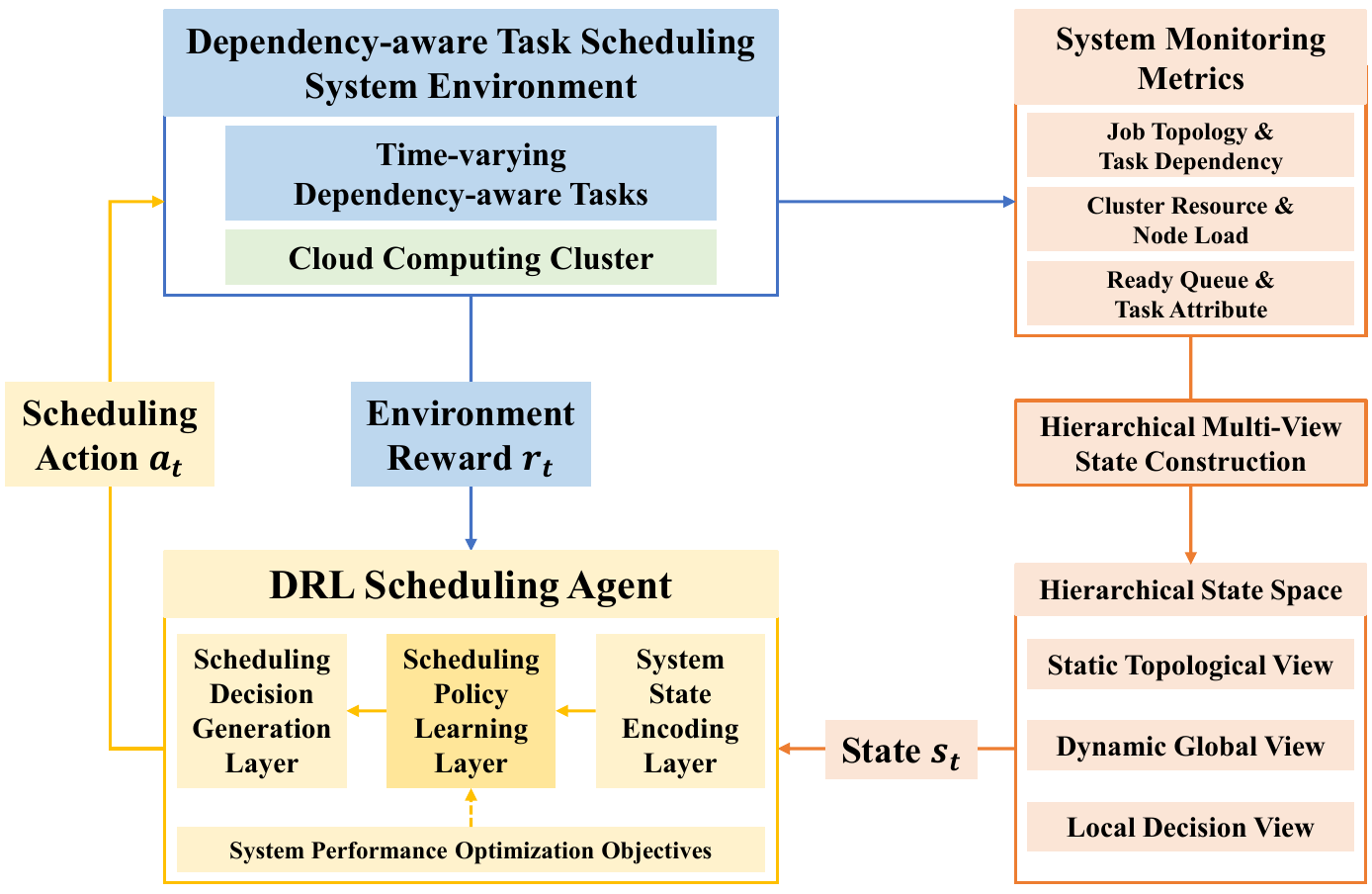}
\caption{Hierarchical Graph Fusion-Driven Reinforcement Learning Model for Dependency-Aware Task Scheduling.}
\label{fig-rlmodel}
\end{figure}

Fig. \ref{fig-rlmodel} illustrates a hierarchical graph fusion–driven RL model for dependency-aware task scheduling.

\noindent {\bf{Hierarchical State Space ($\mathcal{S}$)}}
State representation is crucial to the scheduling performance. For the Job-Task-Instance hierarchy, $s_t \in \mathcal{S}$ is defined as a hierarchical graph fusion state. This hierarchical composite representation integrates three distinct views:

(1) Static Topological View ($\mathcal{H}$): Captures the long-term dependency structure. A hypergraph topology is constructed based on the DAG dependency list, using job ownership identifiers to distinguish subgraphs of different jobs. Node features consist of task static attributes (resource demands and duration) and task topological ranks, characterizing the inherent properties and critical path positions of tasks in the DAG.

(2) Dynamic Global View ($\mathcal{G}_{global}$): Encodes the real-time resource state across the entire cluster. This view aggregates the real-time available resources, machine resource capacities, and machine processing speeds of all computing nodes based on the current system time, forming a physical state snapshot of the cluster. Simultaneously, it combines resource load states with task state vectors and reflects global load pressure using the total number of task instances and the number of pending instances.

(3) Local Decision View ($\mathcal{G}_{local}$): Models the pairwise affinity between the target instance and candidate computing nodes. This view focuses on the current ready task, calculating dynamic waiting duration based on task ready time to measure scheduling urgency. Edge features quantify execution costs by estimating the expected completion time of the task on different nodes, while node features integrate the total number of instances and the number of pending instances for the task.

\noindent {\bf{Action Space ($\mathcal{A}$)}}
The action space is discrete and employs constraint-aware hard masking, $\mathcal{A} = \{1, \dots, |\mathcal{M}|\}$, where action $a_t=k$ represents assigning $T_{target}$ to VM $M_k$. To strictly enforce resource constraints and accelerate exploration, the action space implements hard action masking, defining a binary validity mask vector $\mathbf{m}_t \in \{0, 1\}^{|\mathcal{M}|}$ as shown in Eq. \eqref{eq10}.

\begin{equation}
\label{eq10}
    \mathbf{m}_t[k] = \mathbb{I}(A_k^{cpu}(t) \ge r_{target}^{cpu}) \cdot \mathbb{I}(A_k^{mem}(t) \ge r_{target}^{mem})
\end{equation}

This mask is applied to the unnormalized action scores output by the policy network, effectively constraining the action space to the feasible region $\mathcal{A}_{valid} = \{k \mid \mathbf{m}_t[k]=1\}$. The resource-constraint-based hard action masking mechanism imposes an inductive bias on the model to prevent the agent from learning invalid policies, improving sample efficiency in the early stages of training.

\noindent {\bf{Topology-Prior-Guided Hybrid Reward ($\mathcal{R}$)}}
To address the problem of delayed feedback for long-term objectives in long-horizon scheduling, HiGFRL designs a dense hybrid reward function $R_t$:

\begin{equation}
\label{eq11}
\begin{split}
    R_t = \underbrace{-\alpha_{time} \cdot \delta_{exec}}_{\text{Makespan Minimization}} + \underbrace{\lambda_{eft} \cdot \frac{\min_{k'} \text{EFT}_{k'}}{\text{EFT}_{a_t}}}_{\text{Completion Efficiency}} \\
    + \underbrace{\lambda_{rank} \cdot \text{Rank}(T_{target})}_{\text{Topological Priority}}
\end{split}
\end{equation}

\noindent where $\delta_{exec}$ is the task duration. The Completion Efficiency Incentive term encourages the agent to prefer nodes with earlier completion times in local decisions, thereby improving immediate execution efficiency. Here, $\text{EFT}_{a_t}$ represents the Earliest Finish Time of the task under the selected action $a_t$, and $\min_{k'} \text{EFT}_{k'}$ represents the minimum possible finish time achievable for the task among all feasible VMs. This ratio guides the model towards selecting currently locally optimal resources. The Topological Priority term utilizes the task's upward rank calculated from static graph analysis, i.e., the critical path length from the task to the DAG exit node, as prior knowledge. This enables the agent to identify and prioritize bottleneck tasks on the critical path, thereby optimizing the global job completion time.

\subsection{Multi-View Representation Learning and Fusion Architecture}

\begin{figure*}[!t]
\centering
\includegraphics[width=0.9\textwidth]{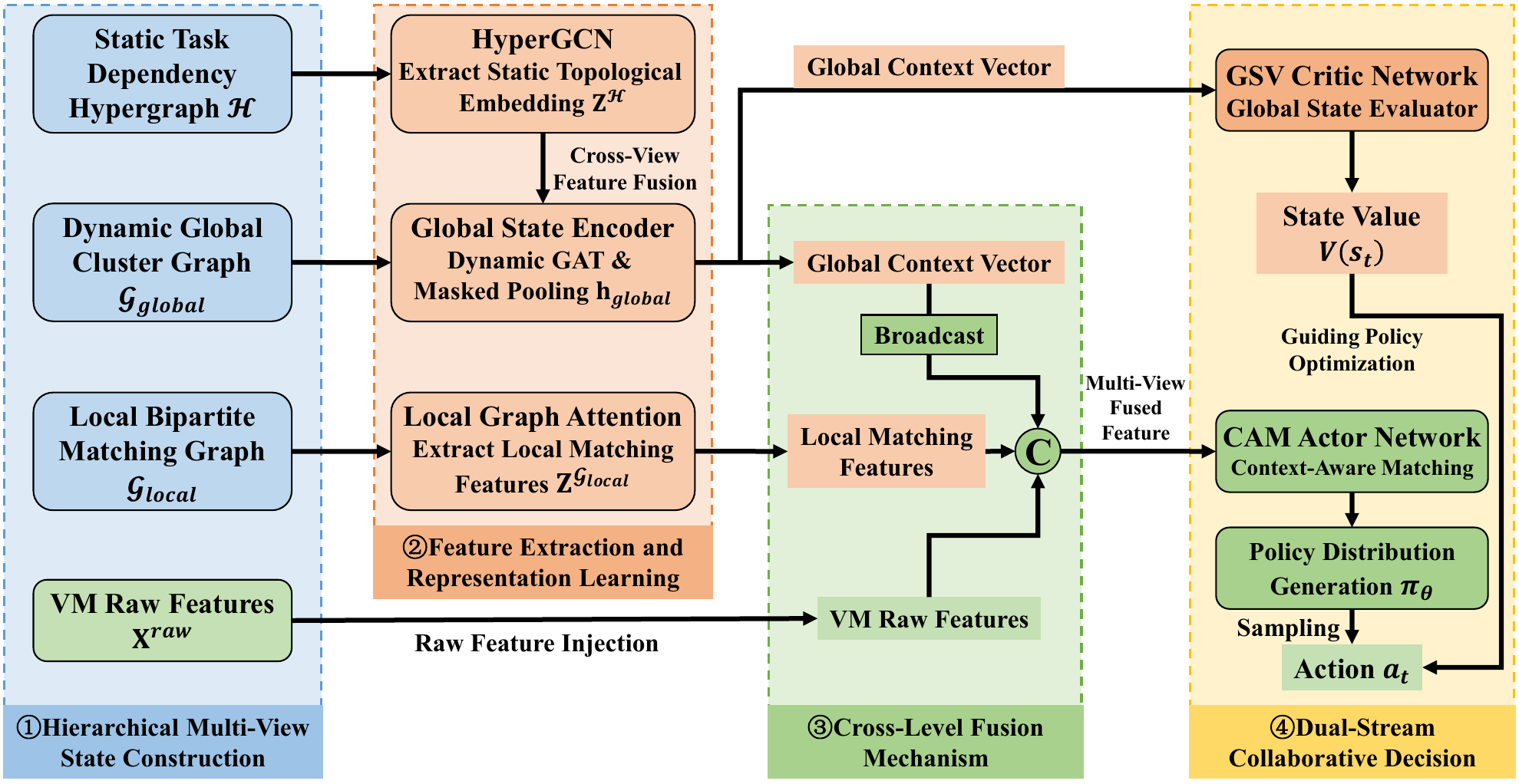}
\caption{Overall Framework of HiGFRL.}
\label{fig-approach}
\end{figure*}

The core challenge of dependency-aware task scheduling lies in effectively modeling the complex interactions between task topological dependencies and machine resource heterogeneity. HiGFRL addresses this issue through a decoupling and fusion strategy, extracting features independently under different views and then performing multi-view fusion through a hierarchical architecture. The overall framework of the HiGFRL approach is illustrated in Fig. \ref{fig-approach}.

{\bf{(1) Hierarchical Graph-Fused State Construction}}

HiGFRL constructs a three-layer graph structure to capture system features at different semantic levels and granularities.

{\bf{Level 1: Static Task Dependency Hypergraph ($\mathcal{H}$).}}
To effectively capture high-order dependencies, HiGFRL models job topology as a hypergraph. Its formal definition is as follows.

\begin{equation}
\label{eq12}
    \mathcal{H} = (\mathcal{V}_{task}, \mathcal{E}_{hyper})
\end{equation}

\noindent where each node $v_i \in \mathcal{V}_{task}$ is characterized by a static feature vector $\mathbf{x}_i^{static}$ containing resource requests, task duration, and topological rank. Explicitly introducing topological rank aims to provide the model with prior knowledge about the task's global position in the DAG.

Predecessor-Aggregation-Based Hyperedge Construction: HiGFRL adopts a Many-to-One dependency aggregation paradigm to construct hyperedges, where each hyperedge $e \in \mathcal{E}_{hyper}$ connects a child task node and all its parent task nodes. This structure explicitly models the synchronization constraint in dependency-aware task scheduling, where a task becomes ready only after all predecessors are completed. Through hypergraph convolution, HiGFRL can directly aggregate features from all predecessors in a single message passing step, thereby efficiently inferring the task's readiness state and avoiding the information dilution problem of ordinary graph convolution when processing long dependency chains.

{\bf{Level 2: Dynamic Global Cluster Graph ($\mathcal{G}_{global}$).}}
This view aims to capture the global real-time state of the heterogeneous cluster. To adapt to the dynamic characteristics of random task flow arrival and completion in dependency-aware task scheduling scenarios, this graph employs a construction strategy combining static topology with dynamic attributes. The formal definition of $\mathcal{G}_{global}$ is as follows.

\begin{equation}
\label{eq13}
    \mathcal{G}_{global} = (\mathcal{V}_{all}, \mathcal{E}_{dep}, \mathbf{M}_{t})
\end{equation}

\noindent where $\mathcal{V}_{all} = \mathcal{V}_{task} \cup \mathcal{V}_{vm}$ covers all task nodes and computing nodes, and $\mathcal{E}_{dep}$ reuses the inherent dependency structure of job DAGs. The dynamic evolution of this view is driven by the following mechanisms:

Real-time Feature Stream Evolution: Node attributes in the graph update in real-time with time step $t$. Task node features $\mathbf{x}^{dyn}$ encode current execution state, resource demands, and completion progress; VM node features $\mathbf{x}^{vm}$ reflect the real-time CPU and memory load rates of each node. This time-varying feature mapping ensures that graph data can characterize the physical state of the environment without latency.

Dynamic Validity Mask ($\mathbf{M}_{t}$): Addressing the dynamic arrival characteristic of jobs, HiGFRL introduces a binary mask vector $\mathbf{M}_{t} \in \{0, 1\}^{|\mathcal{V}_{all}|}$. This mask is dynamically generated by the environment based on task arrival times and completion states. For task nodes that have not yet arrived or have terminated, their corresponding mask values are set to 0. This mechanism allows the model to maintain a fixed graph size while logically filtering out invalid nodes dynamically, providing a precise view of active nodes for subsequent feature aggregation.

{\bf{Level 3: Local Bipartite Matching Graph ($\mathcal{G}_{local}$).}}
This view aims to optimize immediate scheduling decisions through relevance scoring from a local perspective. $\mathcal{G}_{local}$ is defined as a bipartite graph connecting the target task node $u_{target}$ with all candidate VM nodes $\mathcal{V}_{vm}$.

\begin{equation}
\label{eq14}
    \mathcal{G}_{local} = (\{u_{target}\} \cup \mathcal{V}_{vm}, \mathcal{E}_{pair})
\end{equation}

\noindent where the edge $(u_{target}, v_{vm}^k) \in \mathcal{E}_{pair}$ represents a potential assignment path.

Candidate Set Focusing Mechanism: The core purpose of this view is to construct explicit pairwise correlation representations. By explicitly isolating and reinforcing the interaction features between target task demands and specific VM capabilities, this structure forces the model to directly evaluate the affinity of each (Task, VM) matching pair. This not only reduces the interference of global noise on local decisions but also effectively decouples the complexity of action evaluation, ensuring the policy network can capture fine-grained resource matching patterns.

{\bf{(2) Cross-Level Feature Fusion and Dual-Stream Decision}}

HiGFRL proposes a fusion-driven dual-network architecture that hierarchically injects information: from the Static Topological View $\mathcal{H}$ to the Dynamic Global View $\mathcal{G}_{global}$, and finally converging at the Local Decision View $\mathcal{G}_{local}$.

{\bf{Stage 1: Static Topological Feature Extraction}}

First, a hypergraph convolutional network is utilized to process the Static Task Dependency Hypergraph $\mathcal{H} = (\mathcal{V}_{task}, \mathcal{E}_{hyper})$ to capture long-range dependencies and synchronization constraints between tasks. Let $\mathbf{X}^{\mathcal{H}}$ be the node feature matrix. HiGFRL represents hyperedge connections $\mathcal{E}_{hyper}$ as an incidence matrix $\mathbf{H} \in \mathbb{R}^{|\mathcal{V}_{task}| \times |\mathcal{E}_{hyper}|}$, where $H_{ie}=1$ indicates node $v_i$ is contained in hyperedge $e$. The static topological embedding $\mathbf{Z}^{\mathcal{H}}$ is calculated following Eq. \eqref{eq15}.

\begin{equation}
\label{eq15}
    \mathbf{Z}^{\mathcal{H}} = \sigma \left( \mathbf{D}_{\mathcal{V}}^{-\frac{1}{2}} \mathbf{H} \mathbf{W} \mathbf{D}_{\mathcal{E}}^{-1} \mathbf{H}^\top \mathbf{D}_{\mathcal{V}}^{-\frac{1}{2}} \mathbf{X}^{\mathcal{H}} \Theta \right)
\end{equation}

\noindent where $\Theta$ is the learnable filter weight matrix. $\mathbf{D}_{\mathcal{V}}$ and $\mathbf{D}_{\mathcal{E}}$ are diagonal matrices of node degrees and hyperedge degrees, respectively, used for normalization. $\mathbf{W}$ is the hyperedge weight matrix. This formula implements a "node-hyperedge-node" high-order information propagation path via $\mathbf{H} \mathbf{D}_{\mathcal{E}}^{-1} \mathbf{H}^\top$. The generated $\mathbf{Z}^{\mathcal{H}}$ integrates the task's global topological priors and high-order dependency semantics.

{\bf{Stage 2: Global Dynamic Context Aggregation}}

Next, the Global State Encoder is utilized to process the Dynamic Global Cluster Graph $\mathcal{G}_{global}$. In this stage, the model first performs cross-view feature fusion.

Specifically, for task nodes, their features are formed by concatenating the dynamic state $\mathbf{x}^{dyn}$ with the static topological embedding $\mathbf{Z}^{\mathcal{H}}$ extracted in Stage 1; for VM nodes, features are constituted by real-time load features $\mathbf{x}^{vm}$ after dimension alignment. Together, they form the node feature matrix $\mathbf{X}^{\mathcal{G}_{global}}$ of the global graph. This operation injects long-range dependency priors into the real-time state, enhancing the model's ability to anticipate the subsequent impact of tasks.

To capture the complex asymmetric interactions between task urgency and heterogeneous resource loads, the encoder employs a Dynamic Graph Attention mechanism. Taking the fused feature matrix $\mathbf{X}^{\mathcal{G}_{global}}$ as input, it generates the node embedding matrix $\mathbf{H}^{global}$ containing context information.

\begin{equation}
\label{eq16}
    \mathbf{H}^{global} = \text{GAT}_{global}(\mathbf{X}^{\mathcal{G}_{global}}, \mathcal{E}_{dep})
\end{equation}

Subsequently, the encoder uses the mask $\mathbf{M}_{t}$ to perform a masked global pooling operation on the node vectors in $\mathbf{H}^{global}$, generating the global context vector $\mathbf{h}_{global}$ as Eq. \eqref{eq17}.

\begin{equation}
\label{eq17}
    \mathbf{h}_{global} = \frac{\sum_{i=1}^{|\mathcal{V}_{all}|} (\mathbf{M}_{t}[i] \cdot \mathbf{h}_i^{global})}{\sum_{i=1}^{|\mathcal{V}_{all}|} \mathbf{M}_{t}[i] + \epsilon}
\end{equation}

where $\mathbf{h}_i^{global}$ is the row vector of the $i$-th node in matrix $\mathbf{H}^{global}$, representing the high-dimensional feature of node $i$ after attention aggregation; $\epsilon$ is a smoothing term to prevent division by zero. By combining the dynamic graph attention mechanism with masked pooling, this module can dynamically aggregate effective resource load distribution information of the entire cluster while shielding invalid node noise, thereby precisely capturing the global state of the system.

{\bf{Stage 3: Dual-Stream Decision Heads}}

The architecture splits into two specialized streams sharing the global context $\mathbf{h}_{global}$.

Global State Evaluator:
Composed of the Global State-Value Critic network $V_\phi$ (GSV Critic).
This module estimates the value of the current state $V(s_t)$. Taking the global vector $\mathbf{h}_{global}$ as input, it maps the global system state to the expected long-term return via a Multilayer Perceptron (MLP), serving as a baseline for calculating the advantage function. The estimation of $V(s_t)$ is calculated as shown in Eq. \eqref{eq18}.

\begin{equation}
\label{eq18}
    V(s_t) = \text{MLP}_{critic}(\mathbf{h}_{global})
\end{equation}

Context Fusion Allocator:
Composed of the context-aware matching actor network $\pi_\theta$ (CAM Actor). This module executes specific instance-to-node assignments. It operates on the local bipartite graph $\mathcal{G}_{local}$, where $\mathbf{X}^{\mathcal{G}_{local}}$ represents the initial node feature matrix of the graph, containing basic resource attributes and heuristic time estimation features of the target task and candidate VMs. The local graph attention module extracts pairwise local matching features as follows.

\begin{equation}
\label{eq19}
    \mathbf{Z}^{\mathcal{G}_{local}} = \text{GAT}_{local}(\mathbf{X}^{\mathcal{G}_{local}}, \mathcal{E}_{pair})
\end{equation}

To break the limitation of the local view and integrate global context, a cross-level fusion mechanism is introduced. For each candidate VM $M_k$, a multi-view fused feature vector $\mathbf{e}_k^{fused}$ is constructed as follows.

\begin{equation}
\label{eq20}
    \mathbf{e}_k^{fused} = \text{Concat}\left( \mathbf{h}_{k}^{local}, \ \mathcal{B}(\mathbf{h}_{global}), \ \mathbf{x}_{k}^{raw} \right)
\end{equation}

\noindent where $\mathcal{B}(\cdot)$ denotes the broadcast operation, extending the global context vector $\mathbf{h}_{global}$ to match the number of candidate nodes, and $\text{Concat}(\cdot)$ denotes feature dimension concatenation. In this fused representation, $\mathbf{Z}^{\mathcal{G}_{local}}$ represents the feature embedding space of the local bipartite graph, $\mathbf{h}_{k}^{local} \in \mathbf{Z}^{\mathcal{G}_{local}}$ is the local matching feature of the $k$-th candidate VM, aiming to quantify the pairwise correlation between this node and the target task; $\mathbf{h}_{global}$ aligns with the local representation of each candidate VM via the broadcast mechanism, providing unified global resource posture guidance; $\mathbf{x}_{k}^{raw} \in \mathbf{X}^{raw}$ is the raw feature vector of the $k$-th VM directly extracted from the input matrix $\mathbf{X}^{\mathcal{G}_{local}}$ via raw feature injection. This mechanism aims to alleviate the potential over-smoothing problem of deep GNNs, preventing node embeddings from losing precise numerical information after multi-layer aggregation. Since resource scheduling must strictly satisfy resource capacity constraints, explicitly introducing raw features provides the decision layer with a lossless view of node states, ensuring precise perception of remaining node capacity and compliant allocation.

Finally, the fused feature vector $\mathbf{e}_k^{fused}$ is input into an MLP to generate a scalar score $l_k$. This score quantifies the potential value of assigning the current task to VM $M_k$. Subsequently, combining with the validity mask $\mathbf{m}_t$ and applying the Softmax function, the final policy distribution $\pi_\theta(a_t|s_t)$ is calculated as Eq. \eqref{eq21}.

\begin{equation}
\label{eq21}
    l_k = \text{MLP}_{actor}(\mathbf{e}_k^{fused})
\end{equation}

\begin{equation}
\label{eq22}
    \pi_\theta(a_t=k|s_t) = \frac{\mathbf{m}_t[k] \cdot \exp(l_k)}{\sum_{j=1}^{|\mathcal{M}|} \mathbf{m}_t[j] \cdot \exp(l_j)}
\end{equation}

\subsection{Algorithm Training and Policy Optimization}

HiGFRL adopts an end-to-end joint optimization framework based on the Global State Evaluator and Context Fusion Allocator. This framework integrates the aforementioned hybrid GNN architecture, realizing a direct mapping from high-dimensional multi-view graph structures to scheduling decisions. By jointly optimizing the context-aware matching actor network $\pi_\theta$ and the Global State-Value Critic network $V_\phi$ within a shared latent feature space, HiGFRL enables efficient policy iteration in complex heterogeneous environments.

{\bf{(1) Value Baseline Approximation}}

The GSV Critic aims to approximate the value function $V(s_t)$ of the current system state to reduce the variance of policy gradient estimation. Based on the global context vector $\mathbf{h}_{global}$, the GSV Critic outputs a prediction of the cumulative discounted return. Generalized Advantage Estimation (GAE) is used to calculate the advantage function $\hat{A}_t$, which effectively balances bias and variance through a trade-off parameter $\lambda$. The TD error $\delta_t$ is calculated as follows.

\begin{equation}
\label{eq23}
    \delta_t = r_t + \gamma V(s_{t+1}) - V(s_t)
\end{equation}

The advantage estimate $\hat{A}_t$ is the exponentially weighted accumulation of TD errors.

\begin{equation}
\label{eq24}
    \hat{A}_t = \sum_{l=0}^{\infty} (\gamma \lambda)^l \delta_{t+l}
\end{equation}

The parameters $\phi$ of the GSV Critic network are updated by minimizing the mean squared error, ensuring the value baseline accurately reflects the long-term evolution of the global resource posture.

\begin{equation}
\label{eq25}
    L^{Critic}(\phi) = \hat{\mathbb{E}}_t \left[ (V_\phi(s_t) - V_t^{target})^2 \right]
\end{equation}

{\bf{(2) Trust Region Policy Iteration}}

The CAM Actor outputs the action probability distribution $\pi_\theta(a_t|s_t)$ based on the multi-view fused feature $\mathbf{e}_k^{fused}$. To ensure monotonicity and stability of the training process and avoid performance collapse caused by excessively large steps, a clipping mechanism is employed to limit the magnitude of policy updates. Defining the probability ratio of the new policy to the old policy as $r_t(\theta) = \frac{\pi_\theta(a_t|s_t)}{\pi_{\theta_{old}}(a_t|s_t)}$, the policy objective function is designed as Eq. \eqref{eq26}.

\begin{equation}
\label{eq26}
    L^{Actor}(\theta) = \hat{\mathbb{E}}_t \left[ \min \left( r_t(\theta) \hat{A}_t, \ \text{clip}(r_t(\theta), 1-\epsilon, 1+\epsilon) \hat{A}_t \right) \right]
\end{equation}

This objective function constrains policy updates within a trust region $[1-\epsilon, 1+\epsilon]$ via the $\text{clip}(\cdot)$ operation, forcing the model to maintain the stationarity of the policy distribution while optimizing returns.

{\bf{(3) Dynamic Exploration-Exploitation Balance}}

To prevent the model from prematurely converging to suboptimal local optima, an entropy regularization term is introduced into the optimization objective. The entropy of the policy distribution $S[\pi_\theta](s_t)$ serves as part of the reward signal to encourage the agent to maintain action diversity. The total optimization objective combines policy improvement, value approximation, and entropy maximization as follows.

\begin{equation}
\label{eq27}
    L(\theta, \phi) = \hat{\mathbb{E}}_t \left[ L^{Actor}(\theta) - c_1 L^{Critic}(\phi) + \beta_t S[\pi_\theta](s_t) \right]
\end{equation}

\noindent where the entropy coefficient $\beta_t$ follows a linear decay schedule.

\begin{equation}
\label{eq28}
    \beta_t = \max(\beta_{end}, \beta_{start} - \frac{t}{T_{max}}(\beta_{start} - \beta_{end}))
\end{equation}

This dynamic adjustment mechanism enables the agent to possess higher entropy weight in the initial training stage for broad exploration of the solution space, and gradually reduce the weight in later stages to focus on refined exploitation of high-value regions, thereby achieving an adaptive balance between exploration and exploitation.

\section{Performance Evaluation}

HiGFRL is implemented based on Python 3.8.18 and PyTorch 1.13.1 with CUDA support, and its GNN components are implemented via PyTorch Geometric 2.0.4. All experiments are conducted on a cloud server equipped with a 20-core CPU, 80 GB of system memory, and an NVIDIA GeForce RTX 4090 GPU. This section presents experimental results and analyses aiming to answer the following research questions. The source code and datasets of HiGFRL are available at https://github.com/igeng/HiGFRL.

\begin{itemize}

\item{\textbf{RQ1 (Training Convergence Performance:} How does the convergence efficiency of HiGFRL compare to other DRL-based baselines during the training phase, and does it demonstrate superior stability?}

\item{\textbf{RQ2 (Scheduling Effectiveness):} How effective is HiGFRL in optimizing overall scheduling performance, particularly in minimizing Makespan and Average Task Flow Time, compared to heuristics and DRL-based approaches under different workload scales?}

\item{\textbf{RQ3 (System Operational Efficiency):} How does HiGFRL perform in optimizing system operational efficiency under different workload scales, specifically in terms of minimizing scheduling overheads including Total Retry Counts and Average Task Wait Times?}

\item{\textbf{RQ4 (Ablation Analysis):} How does the hierarchical graph fusion mechanism contribute to the overall system performance, and what is its necessity?}

\end{itemize}

\subsection{Experimental Setup}

{\bf{Cluster Resource Configuration.}} In the training phase, the experiment sets up a cloud computing cluster consisting of 40 VMs. As summarized in Table~\ref{tab:table3}, four VM types are considered, each with different resource configurations.

\begin{table}[!t]
\caption{CLUSTER RESOURCE CONFIGURATION DETAILS\label{tab:table3}}
\centering
\begin{tabular}{c c c c c}
\hline
Types & CPU cores & Memory capacity & Quantity & Processing speed\\
\hline
1 & 2 & 4 & 10 & 2\\
2 & 4 & 8 & 10 & 4\\
3 & 8 & 16 & 10 & 8\\
4 & 16 & 32 & 10 & 16\\
\hline
\end{tabular}
\end{table}

{\bf{Dependency–aware Task Workload Patterns.}} 
Experiments are based on the Alibaba Cluster Trace v2018 dataset, which contains real production workload traces collected from approximately 4000 machines over 8 days. Tasks with explicit DAG structures are selected from batch job workloads to evaluate dependency-aware task scheduling and policy learning performance. Specifically, jobs are first sampled from the workload trace dataset, consisting of tasks of varying sizes, to form a training set containing a total of 1000 jobs and 4076 tasks. Subsequently, to further evaluate the adaptability of the approach, three scales of workload patterns are designed, with two different maximum instance configurations for each pattern, resulting in six experimental scenarios as follows:

\begin{itemize}
\item{\emph{Standard-scale pattern.}
300 jobs, 2160 tasks, maximum instance counts of 50 and 100, referred to as Standard-50 and Standard-100.}

\item{\emph{Medium-scale pattern.}
600 jobs, 4417 tasks, maximum instance counts of 50 and 100, referred to as Medium-50 and Medium-100.}

\item{\emph{Large-scale pattern.}
900 jobs, 6704 tasks, maximum instance counts of 100 and 250, referred to as Large-100 and Large-250.}
\end{itemize}

In the experiments, the six experimental configurations will be directly referenced using the above abbreviations for ease of description and comparison.

{\bf{Baselines.}} We evaluate HiGFRL against six approaches, including three classic heuristics approaches and three advanced approaches combining GNNs and DRL:

\begin{itemize}
\item{\emph{Random.}
Assigns tasks to a randomly selected VM from the set of currently feasible nodes.}

\item{\emph{Round Robin.}
Allocates tasks to VMs in a fixed cyclic order, leaving tasks queued for future retries if the target VM lacks sufficient resources.}

\item{\emph{Greedy-Tetris.}
A multi-dimensional bin-packing heuristic that selects the VM maximizing the dot product between task resource demands and available VM resources to minimize resource  fragmentation \cite{ref2-23}.}

\item{\emph{READYS.}
A representative Actor-Critic approach utilizing GCNs for DAG topological feature extraction, applying global pooling to aggregate node embeddings into a graph-level state \cite{ref2-13}.}

\item{\emph{GA-DRL.}
A value-based DRL-based approach integrating Bi-directional GATs with Double DQN, aggregating predecessor and successor information to capture complex DAG dependencies for placement decisions \cite{ref2-11}.}

\item{\emph{MODRL.} 
A D3QN-based hybrid framework utilizing GAT, LSTM, and Set Transformer to respectively capture topological, sequential, and cluster features, enhanced by prioritized experience replay for stable multi-objective optimization \cite{ref2-12}.}

\end{itemize}

{\bf{Evaluation metrics.}} 
To comprehensively evaluate the performance of HiGFRL against baselines in complex scheduling scenarios, four metrics are selected: Makespan, Average Task Flow Time, Average Task Wait Time, and Total Retry Count. These metrics comprehensively measure the system from the dimensions of efficiency, responsiveness, and scheduling stability.

(1) Makespan

Makespan represents the total time required to complete the execution of the entire workload. It is defined as the completion time of the last finished task in the system minus the start time. A smaller Makespan indicates higher throughput and better overall efficiency of the cluster. Let $\mathcal{T}$ be the set of all completed tasks, and $C_i$ be the completion time of task $i$. The Makespan is mathematically formulated as follows.

\begin{equation}
    MK = \max_{i \in \mathcal{T}} (C_i) - T_{start}
\end{equation}

\noindent where $T_{start}$ represents the task arrival start time.

(2) ATFT (Average Task Flow Time)

Task flow time measures the total duration a task spends in the system, covering waiting time in the ready queue and actual execution time on the virtual machine, reflecting system responsiveness from the user's perspective. ATFT is calculated by averaging the flow times of all successfully completed tasks. Let $A_i$ be the arrival time of task $i$. ATFT is defined as Eq. \eqref{eq30}.

\begin{equation}
\label{eq30}
    ATFT = \frac{1}{|\mathcal{T}|} \sum_{i \in \mathcal{T}} (C_i - A_i)
\end{equation}

\noindent where $|\mathcal{T}|$ represents the total number of completed tasks. A lower ATFT implies tasks are processed and completed faster after submission.

(3) ATWT (Average Task Wait Time)

To further analyze the source of latency, ATWT is used to measure the scheduling delay. It is defined as the time interval between the task arrival moment and the moment the scheduler first attempts to assign it to a VM. Unlike flow time, this metric specifically quantifies queuing delays caused by dependency constraints or resource contention before task execution. Let $S^{first}_i$ be the timestamp of the first scheduling attempt for task $i$. The ATWT is calculated as follows. Minimizing ATWT implies that the scheduler is efficient in handling pending tasks and reducing queue congestion.

\begin{equation}
    ATWT = \frac{1}{|\mathcal{T}|} \sum_{i \in \mathcal{T}} (S^{first}_i - A_i)
\end{equation} 

(4) TRC (Total Retry Count)

In a resource-constrained cluster, a scheduling decision may fail if the selected VM does not have sufficient available CPU or Memory at the specific moment of assignment. When this occurs, the task is rejected and must be re-queued for a later attempt. The TRC aggregates the number of such failed scheduling attempts across the entire scheduling process. Let $R_i$ be the number of failed attempts for task $i$. The TRC is defined as follows.

\begin{equation}
    TRC = \sum_{i \in \mathcal{T}_{all}} R_i
\end{equation}

\noindent where $\mathcal{T}_{all}$ represents all tasks in the workload. A lower TRC indicates that the approach is making more accurate decisions by effectively respecting resource constraints, thereby reducing scheduling overhead and wasted computational cycles.

\subsection{Training Convergence Performance Analysis}

\begin{figure}[!t]
\centering
\includegraphics[width=0.45\textwidth]{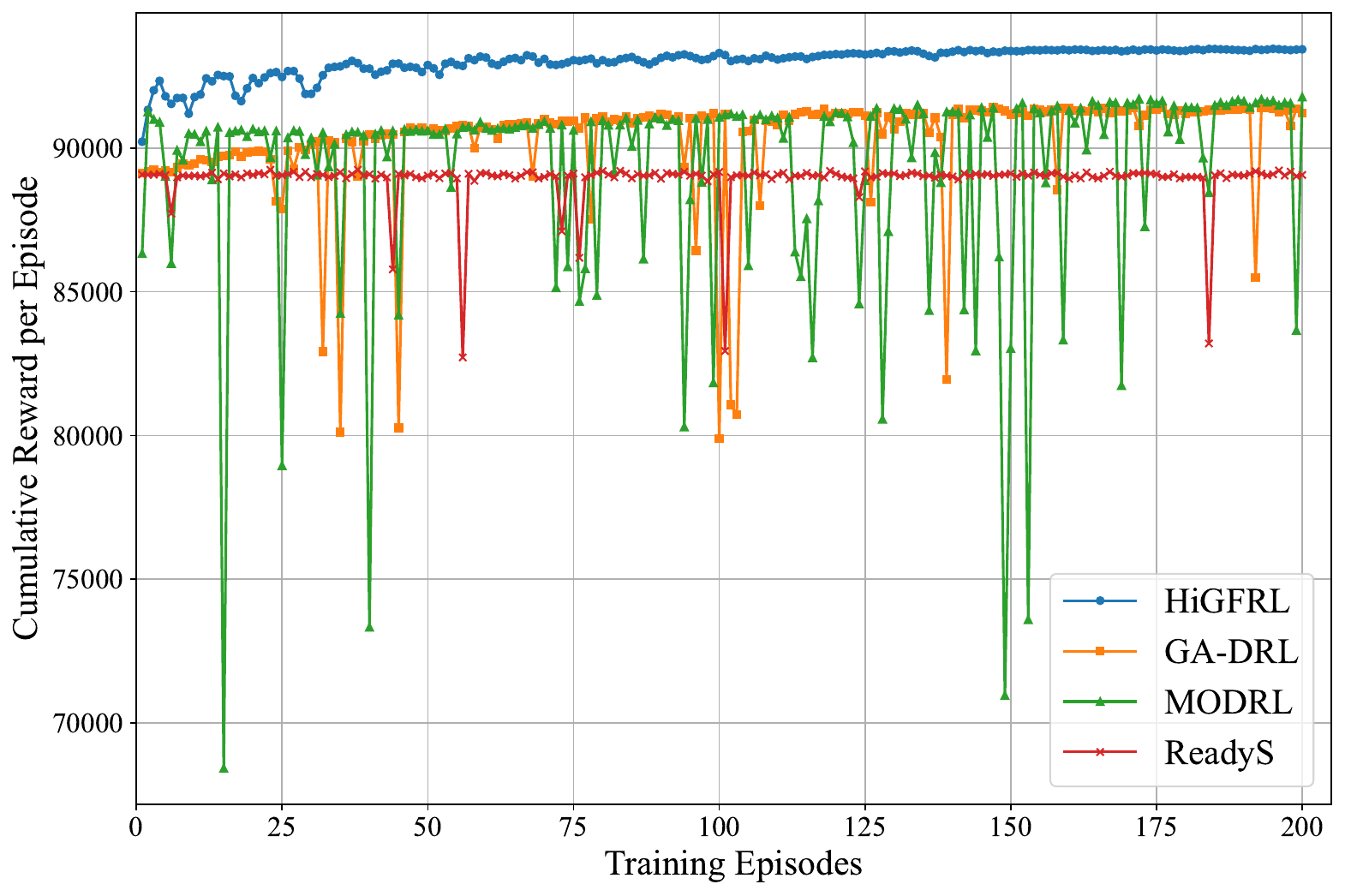}
\caption{Comparison of cumulative reward per episode convergence curves across DRL-based approaches.}
\label{RL_conver}
\end{figure}

Fig. \ref{RL_conver} presents a comparative analysis of the cumulative reward per episode convergence curves for HiGFRL and three DRL-based baselines: GA-DRL, MODRL, and READYS. As illustrated, HiGFRL exhibits superior convergence efficiency and asymptotic stability throughout the training phase. Starting with a robust initial performance, HiGFRL demonstrates a rapid ascent within the first 20 episodes, quickly stabilizing at a high-level convergence value of approximately 93400.

In contrast, the baselines struggle to achieve comparable optimality, highlighting significant deficiencies in how they integrate resource heterogeneity into their state representations. GA-DRL shows a steady but slow improvement, eventually plateauing around 91400. While its Bi-GAT mechanism effectively captures task dependencies, the architecture structurally treats resource states as isolated feature vectors rather than integrating them into the graph structure. This limits the agent's ability to learn complex Task-VM affinities within the graph context. Furthermore, its $\epsilon$-greedy policy restricts exploration efficiency in high-dimensional action spaces, contributing to its slower convergence rate. 

MODRL exhibits significant numerical oscillation and suboptimal convergence trends, fluctuating significantly between 70000 and 90000 during the early stages of training. This instability is partly attributed to the high variance characteristics of value-based approaches in highly stochastic environments. More critically, its graph embedding approach tends to over-compress resource information into a single global embedding, obscuring the fine-grained availability details of heterogeneous nodes and limiting the precision of scheduling decisions. In the later stages of training, MODRL converges to a level similar to that of GA-DRL. Meanwhile, READYS displays early performance saturation, hovering around 89000 with limited policy improvement in subsequent stages. This suggests that its basic GCN architecture focuses primarily on topological encoding, and its combination with the actor-critic mechanism often treats resources as static attributes rather than interactive graph nodes, failing to effectively capture the fine-grained interactions between complex task dependencies and heterogeneous resources.

The superior performance of HiGFRL strongly validates the effectiveness of its hierarchical graph fusion architecture. Unlike baselines that tend to marginalize or over-compress resource information during graph construction, HiGFRL constructs a multi-view state representation that explicitly fuses task topology and resource heterogeneity at both global and local levels. This precise structured information and semantic context enable the agent to make more informed scheduling decisions. Additionally, the synergy between this architecture and the linear entropy decay strategy ensures a dynamic balance between broad exploration and stable exploitation, effectively preventing premature policy convergence and helping the agent explore better scheduling policies.

\subsection{Overall Scheduling Effectiveness}

\begin{figure}[!t]
\centering
\includegraphics[width=0.45\textwidth]{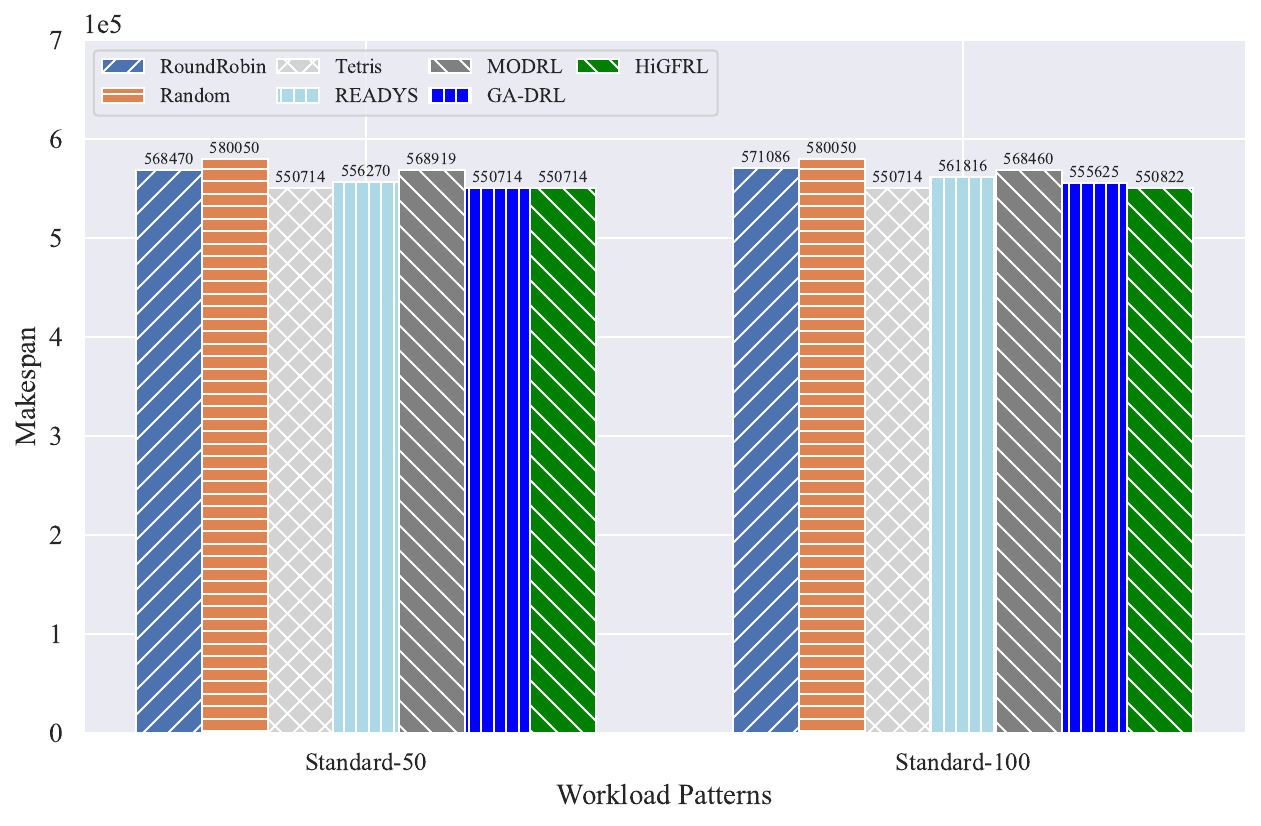}
\caption{Makespan comparison of different approaches under standard-scale workload patterns.}
\label{makespan-1}
\end{figure}

\begin{figure}[!t]
\centering
\includegraphics[width=0.45\textwidth]{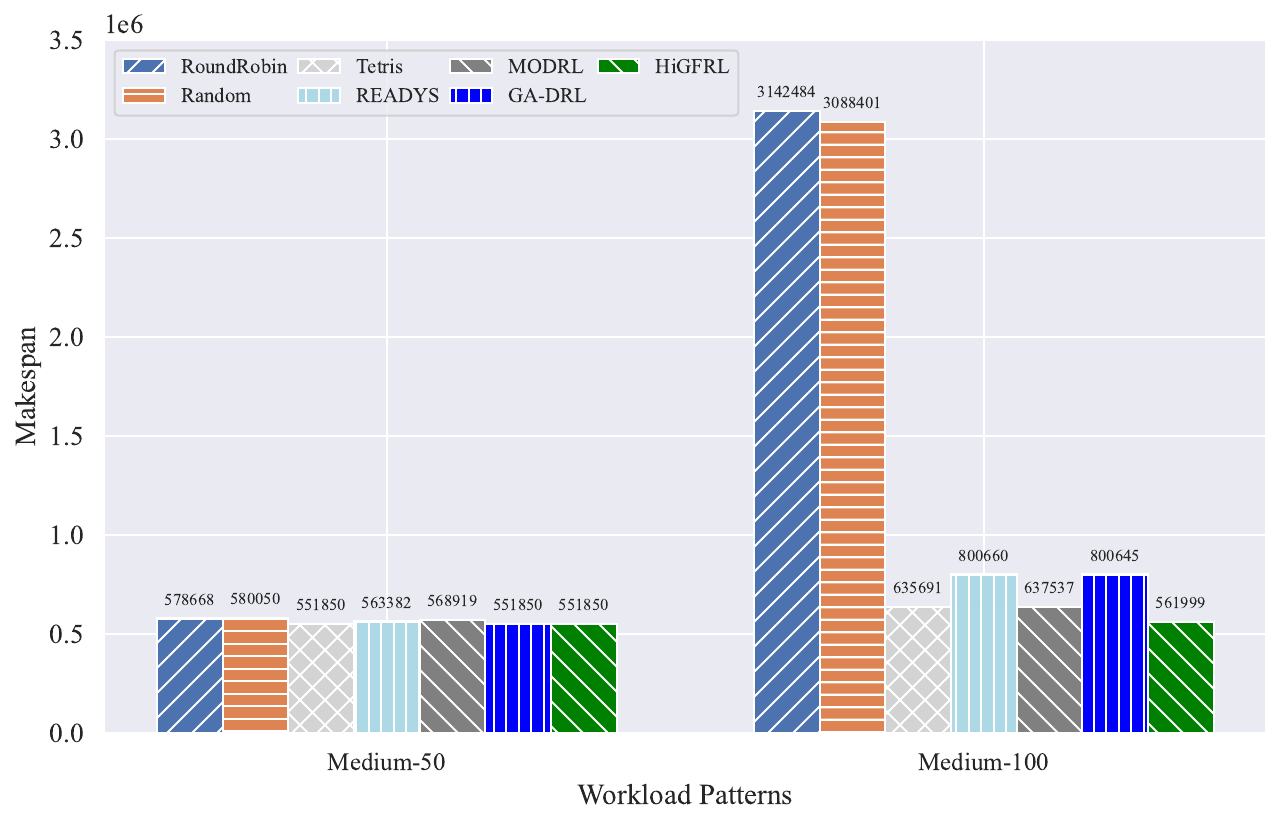}
\caption{Makespan comparison of different approaches under medium-scale workload patterns.}
\label{makespan-2}
\end{figure}

\begin{figure}[!t]
\centering
\includegraphics[width=0.45\textwidth]{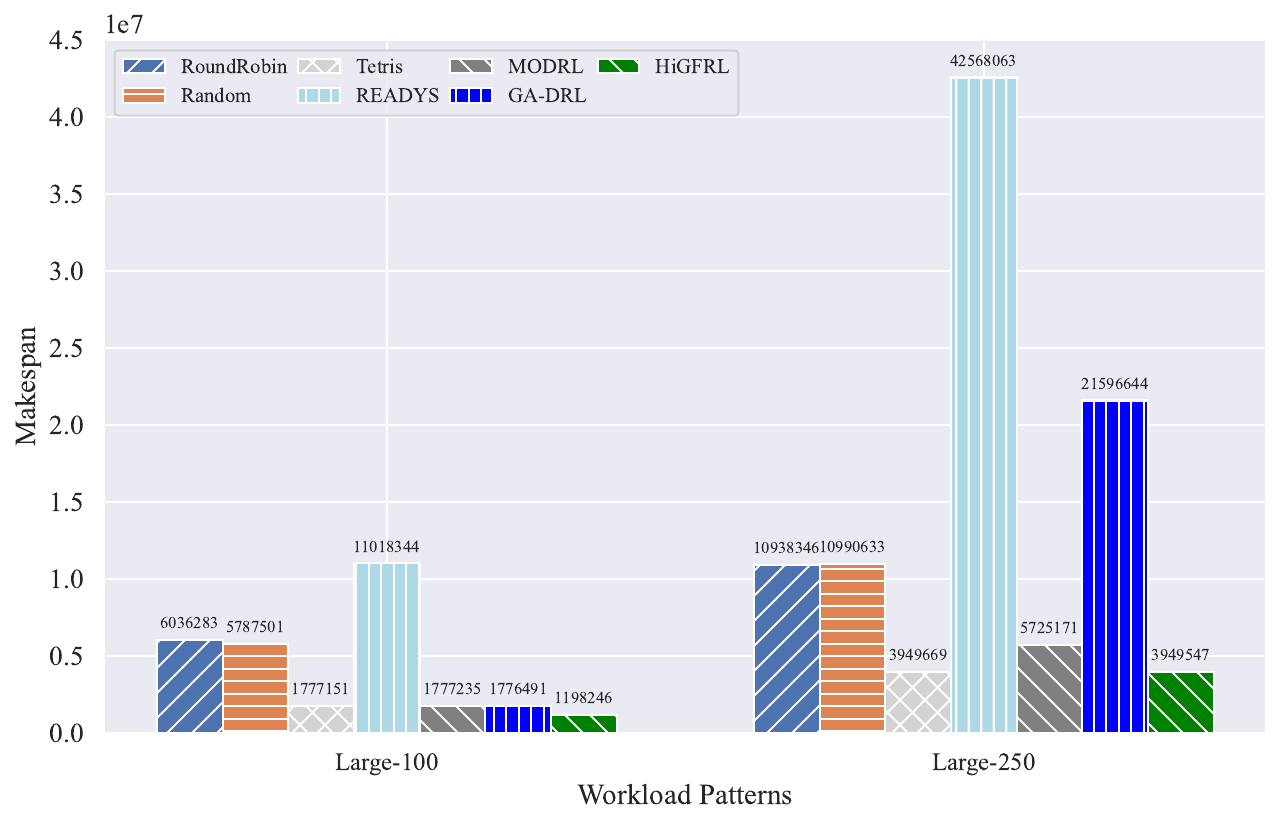}
\caption{Makespan comparison of different approaches under large-scale workload patterns.}
\label{makespan-3}
\end{figure}

\begin{figure}[!t]
\centering
\includegraphics[width=0.45\textwidth]{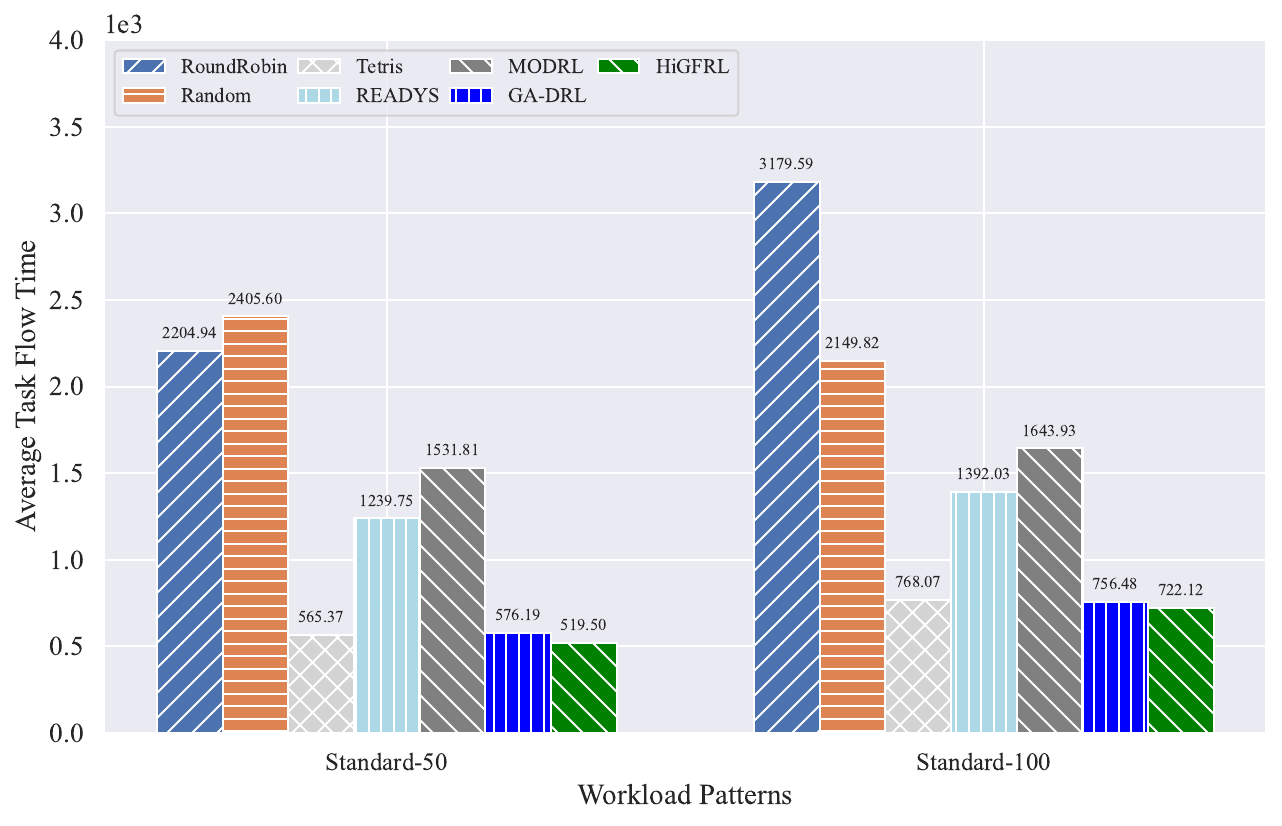}
\caption{Average task flow time comparison of different approaches under standard-scale workload patterns.}
\label{atft-1}
\end{figure}

\begin{figure}[!t]
\centering
\includegraphics[width=0.45\textwidth]{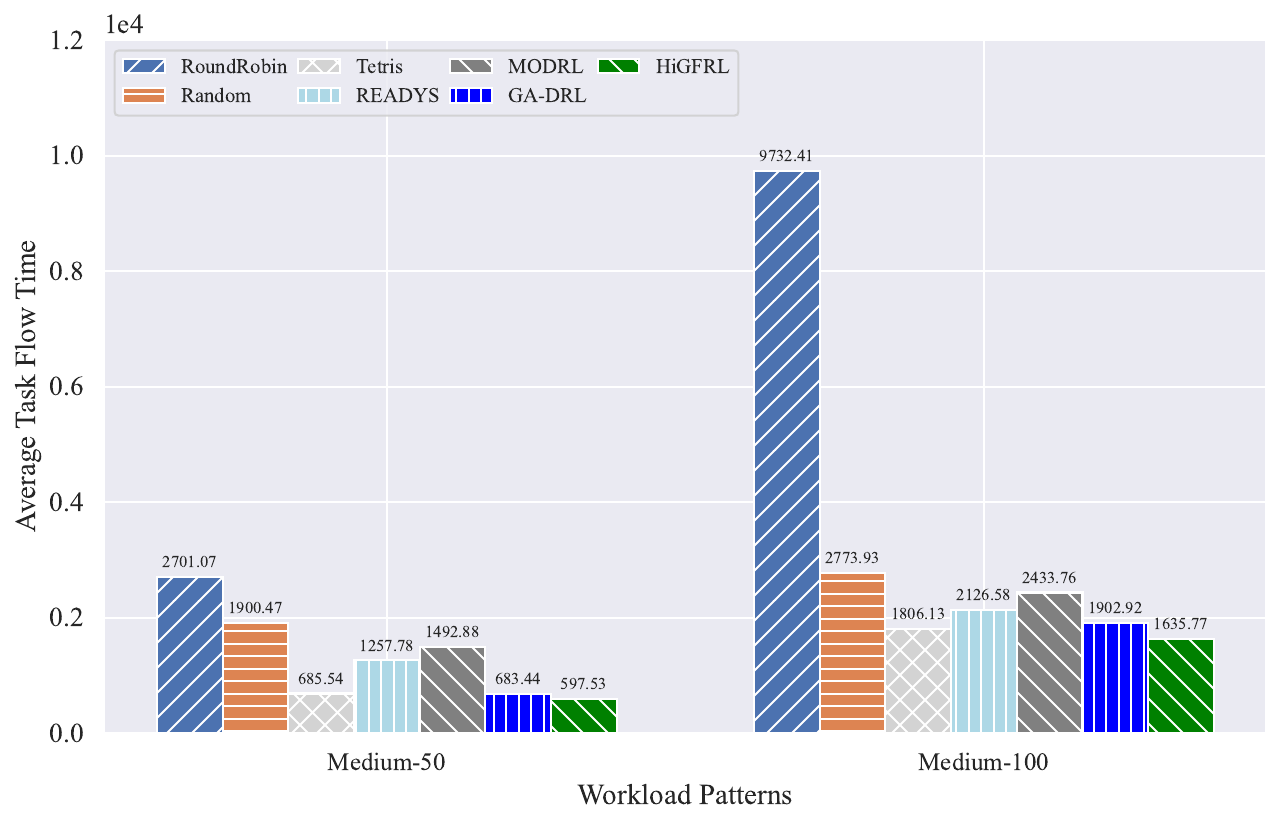}
\caption{Average task flow time comparison of different approaches under medium-scale workload patterns.}
\label{atft-2}
\end{figure}

\begin{figure}[!t]
\centering
\includegraphics[width=0.45\textwidth]{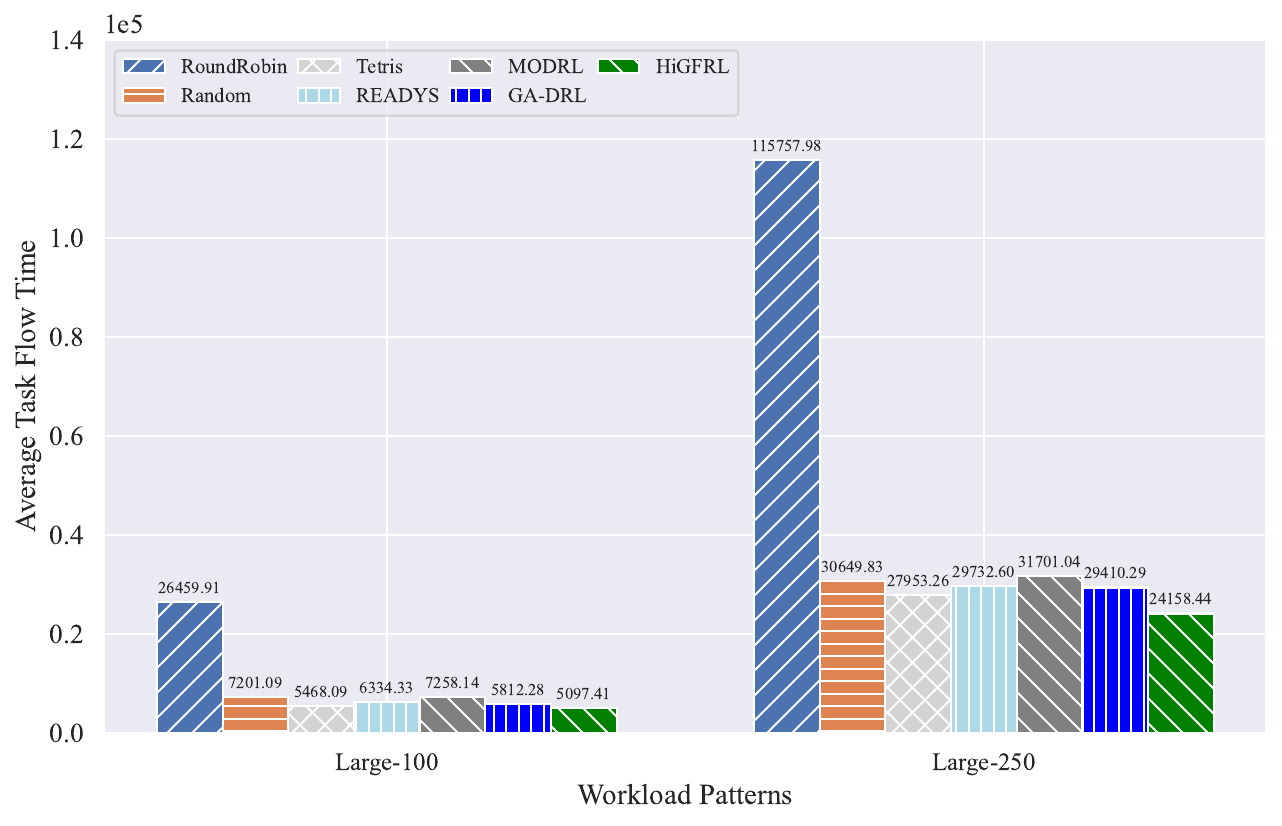}
\caption{Average task flow time comparison of different approaches under large-scale workload patterns.}
\label{atft-3}
\end{figure}

To rigorously evaluate the scheduling efficiency of HiGFRL relative to the six baseline approaches, the performance of different approaches on Makespan and ATFT is analyzed. Fig. \ref{makespan-1} to Fig. \ref{makespan-3} present the comparative results for Makespan, while Fig. \ref{atft-1} to Fig. \ref{atft-3} illustrate the results for ATFT under three different scales of workload patterns, respectively. Experimental results indicate that HiGFRL achieves optimal or near-optimal scheduling performance in all test scenarios, and its performance advantage significantly expands as workload scale and scheduling difficulty increase.

In the standard-scale scenarios of Standard-50 and Standard-100, HiGFRL exhibits extremely stable performance in the Makespan metric, with values of 550714 and 550822, respectively. These results are on par with the best-performing heuristic approach, Greedy-Tetris, and GA-DRL, while significantly outperforming READYS at 561816 and MODRL at 568460. However, Makespan alone does not fully reflect differences in scheduling quality. Observing the ATFT metric in Fig. \ref{atft-1}, it is evident that even when the Makespan gap is narrow, HiGFRL maintains a lead in service quality. For instance, in Standard-50, HiGFRL's average flow time is 519.50, distinctly lower than Greedy-Tetris's 565.37 and GA-DRL's 576.19. This indicates that while several strong baselines can complete the entire workload in a similar timeframe, HiGFRL prioritizes critical tasks more effectively through superior resource matching, thereby reducing the average residence time of tasks in the system.

As the workload scale expands to medium levels, the advantages of HiGFRL become more pronounced. As shown in Fig. \ref{makespan-2}, in the high-concurrency Medium-100 scenario, basic approaches like RoundRobin and Random suffer a performance collapse due to frequent resource conflicts, with Makespan soaring above 3000000. At this point, GA-DRL also exhibits instability, reaching a Makespan of 800645. In contrast, HiGFRL maintains the Makespan at a low level of 561999, significantly outperforming MODRL's 637537. Regarding the ATFT metric in Fig. \ref{atft-2}, HiGFRL achieves an outstanding result of 1635.77 in the Medium-100 scenario, which is 9.43\% lower than the second-best Greedy-Tetris and 14.04\% lower than GA-DRL. This widening gap is primarily attributed to the hierarchical graph fusion architecture of HiGFRL. Unlike GA-DRL and READYS, which are prone to feature over-smoothing on larger scale graphs, HiGFRL explicitly models high-order dependencies via hypergraph convolution and preserves feature differences of heterogeneous resources under a global view. This allows it to accurately identify critical paths affecting overall scheduling even as task volume surges.

In large-scale scenarios, HiGFRL demonstrates decisive performance dominance. As illustrated in Fig. \ref{makespan-3}, in the most complex Large-100 scenario, HiGFRL's Makespan is only 1198246, whereas the closest competitors, GA-DRL and Greedy-Tetris, reach values as high as 1776491 and 1777151, respectively. HiGFRL achieves a 32.55\% reduction in Makespan in this scenario. More notably, regarding the QoS metrics shown in Fig. \ref{atft-3}, under the extreme load configuration of Large-250, HiGFRL controls the ATFT at 24158.44, while all other comparison approaches exceed 27000. In this scenario, HiGFRL achieves a 13.58\% performance improvement compared to the second-best MODRL, 27953.26. This result fully proves the robustness of HiGFRL under extreme pressure. This is mainly attributed to HiGFRL's unique fusion-driven dual-stream network design, where the global state evaluator, based on joint encoding of static topology and dynamic load, provides precise long-term value prediction, guiding the context fusion allocator to perform efficient policy iteration in a massive action space. Coupled with the adaptive entropy regularization mechanism, this architecture effectively avoids premature policy convergence in large-scale state spaces. Consequently, it maintains a globally optimal decision-making vision throughout long scheduling sequences, overcoming the limitations of other graph RL baselines that are prone to falling into local optima in complex scenarios.

\subsection{Operational Efficiency and Stability Analysis}

\begin{figure}[!t]
\centering
\includegraphics[width=0.45\textwidth]{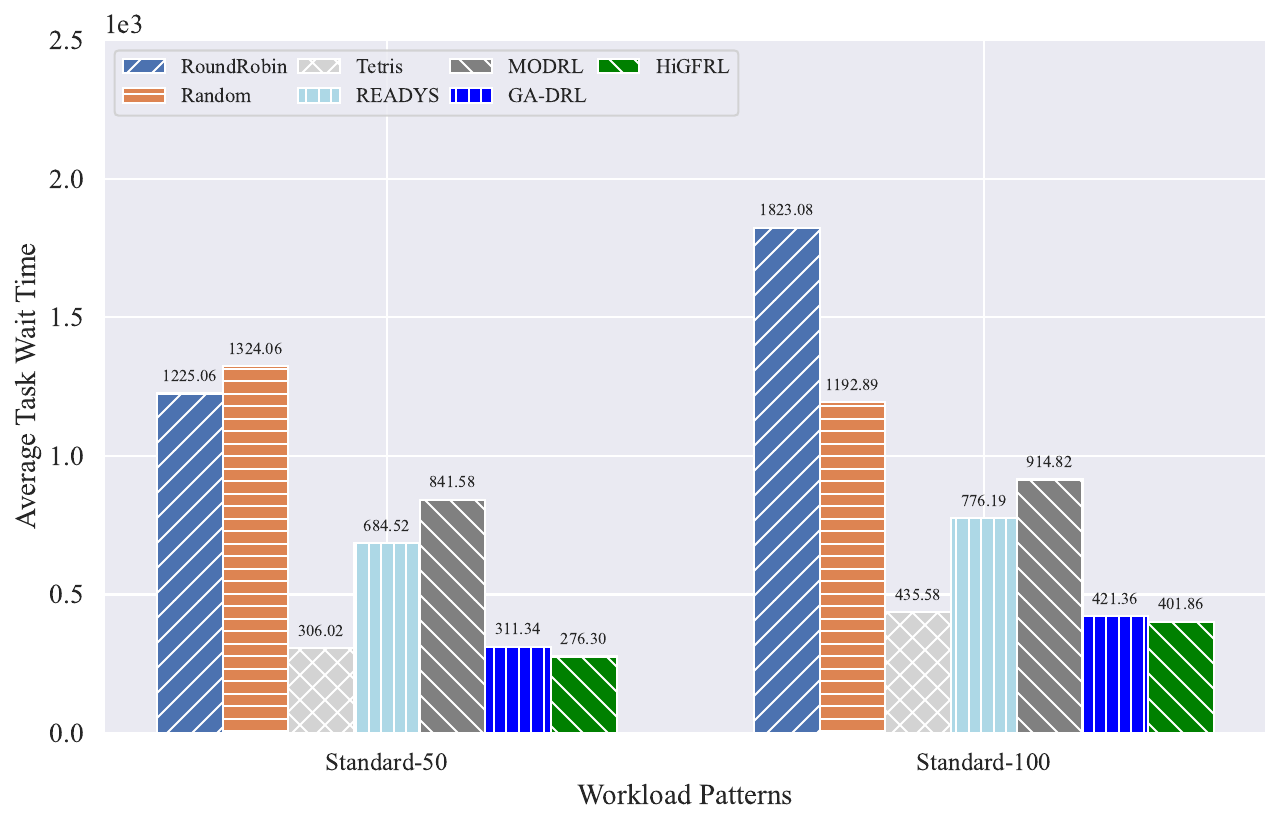}
\caption{Average task wait time comparison of different approaches under standard-scale workload patterns.}
\label{atwt-1}
\end{figure}

\begin{figure}[!t]
\centering
\includegraphics[width=0.45\textwidth]{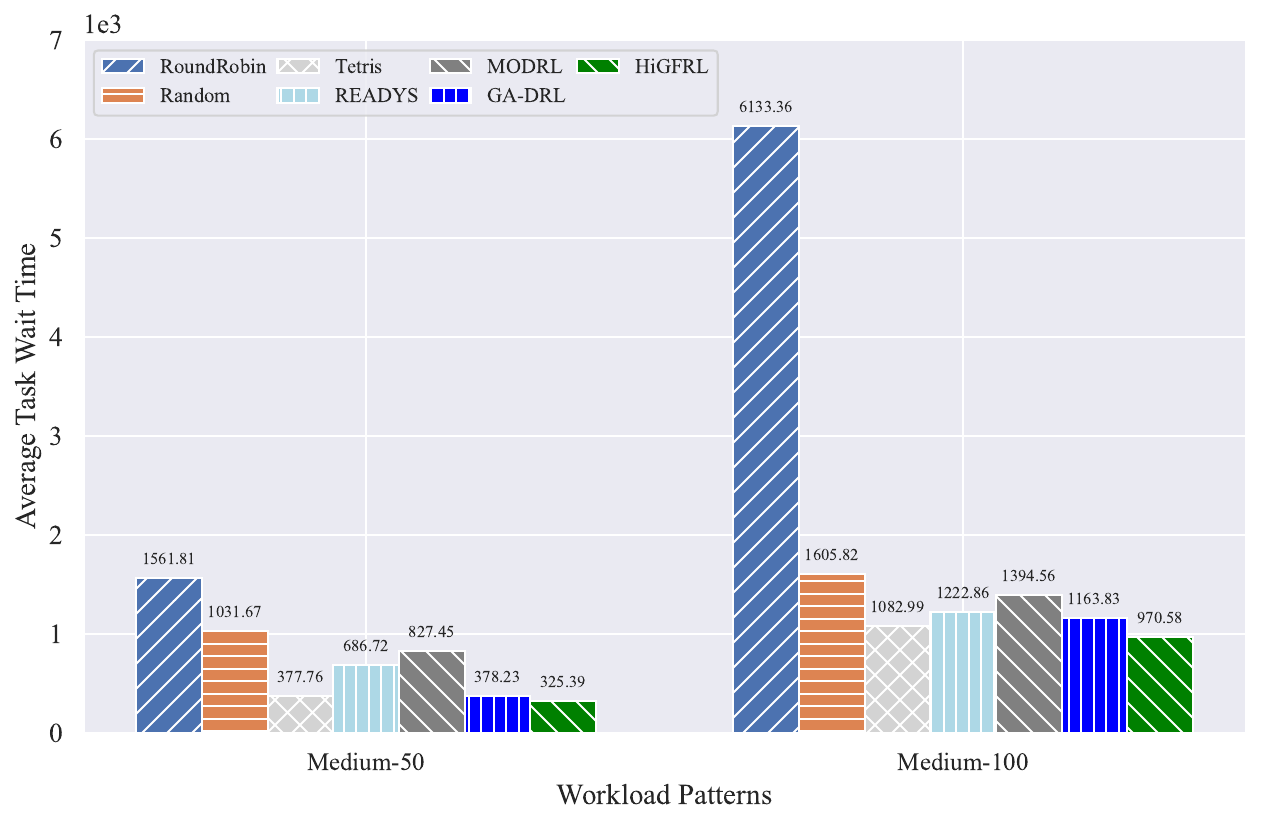}
\caption{Average task wait time comparison of different approaches under medium-scale workload patterns.}
\label{atwt-2}
\end{figure}

\begin{figure}[!t]
\centering
\includegraphics[width=0.45\textwidth]{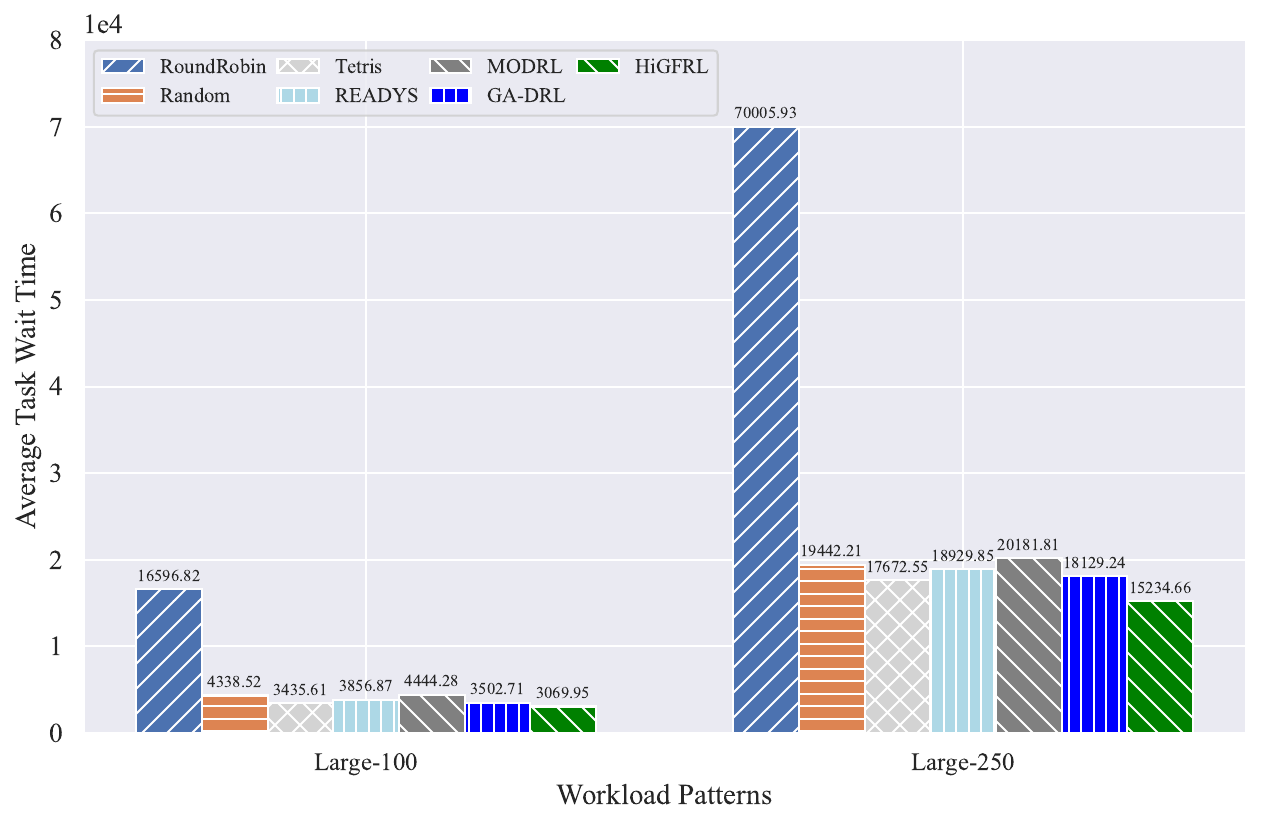}
\caption{Average task wait time comparison of different approaches under large-scale workload patterns.}
\label{atwt-3}
\end{figure}

To thoroughly investigate the operational efficiency and stability of the system under varying workloads, we further analyzed the ATWT and TRC. These two metrics directly reflect queue congestion levels and the precision of resource matching. The comparative results for ATWT are shown in Fig. \ref{atwt-1} to Fig. \ref{atwt-3}, and the statistics for TRC are presented in Table \ref{tab:trc_comparison}. Experimental data consistently indicate that HiGFRL has a significant advantage in optimizing operational efficiency. 

In terms of the ATWT metric, HiGFRL maintains the lowest record across all six experimental configurations, with the margin of superiority expanding as the load increases. Specifically, in the Standard-50 scenario, HiGFRL compresses the average wait time to 276.30, superior to 306.02 achieved by Tetris and 311.34 by GA-DRL. When the scenario switches to Large-100, HiGFRL records an ATWT of only 3069.95, representing a reduction in queuing latency of hundreds of seconds compared to GA-DRL at 3502.71 and READYS at 3856.87. This low-latency characteristic benefits primarily from the combination of HiGFRL's global dynamic state encoder and local bipartite graph. Unlike READYS which uses GCNs for indiscriminate average aggregation of neighbor node features, HiGFRL's local graph explicitly encodes interaction features between tasks and specific VMs, combined with dynamic weighting via attention mechanisms. This allows the scheduler to capture the dynamic distribution of resource fragments and available time windows in the cluster in real-time, thereby rapidly dispatching tasks from the ready queue and greatly alleviating queue backlog. In the most challenging Large-250 scenario in Fig. \ref{atwt-3}, the gap is particularly evident. HiGFRL's average wait time is 15234.66, while the best-performing baseline algorithm, Tetris, is 17672.55. HiGFRL reduces wait time by 13.79\%, directly attributed to its local bipartite graph matching mechanism's acute capture of immediate resource fragments.

Regarding scheduling stability, the TRC reflects the approach's adherence to resource constraints. As shown in Table \ref{tab:trc_comparison}, HiGFRL achieves the lowest retry counts in the majority of high-load scenarios, including Standard-100, Medium-50, Medium-100, and Large-250. Notably, in the most challenging Large-250 scenario, HiGFRL limits its retry count to 11192, a figure significantly better than the 11878 retries of GA-DRL and 11690 of MODRL, demonstrating superior resource adaptability. It is worth noting that in specific scenarios such as Standard-50 and Large-100, HiGFRL's retry count is marginally higher than that of Tetris or GA-DRL. For instance, in the Large-100 scenario, HiGFRL records 8523 retries compared to 8147 for GA-DRL. However, considering the ATFT and ATWT data, although HiGFRL incurs a minimal number of additional retries, its corresponding flow and wait times are significantly lower. This indicates that HiGFRL has learned a more advanced policy: it does not mechanically pursue zero retries but rather engages in policy exploration for superior resource allocation. This mechanism allows the approach to skip current suboptimal solutions with minimal retry costs in exchange for substantial improvements in global scheduling efficiency and user service quality, thereby achieving an optimal balance among multiple key evaluation metrics.

\begin{table*}[!t]
  \centering
  \caption{Comparison of Total Retry Count under different workload patterns}
  \label{tab:trc_comparison}
  \begin{tabular}{p{1.2in}<{\centering}p{0.7in}<{\centering}p{0.6in}<{\centering}p{0.6in}<{\centering}p{0.6in}<{\centering}p{0.6in}<{\centering}p{0.6in}<{\centering}p{0.6in}<{\centering}}
    \hline
    \textbf{Workload Patterns} & \textbf{RoundRobin} & \textbf{Random} & \textbf{Tetris} & \textbf{READYS} & \textbf{MODRL} & \textbf{GA-DRL} & \textbf{HiGFRL} \\
    \hline
    Standard-50  & 3301  & 21    & \textbf{0}     & 32    & 7    & 5     & 13 \\
    Standard-100 & 4962  & 569   & 435   & 439   & 519   & 513   & \textbf{358} \\
    Medium-50    & 9890  & 967   & 881   & 842   & 996   & 584   & \textbf{578} \\
    Medium-100   & 13798 & 3388  & 2702  & 2760  & 3051  & 2870  & \textbf{2559} \\
    Large-100    & 26200 & 9143  & 8530  & 9053  & 9067  & \textbf{8147}  & 8523 \\
    Large-250    & 30263 & 12155 & 11540 & 11450 & 11690 & 11878 & \textbf{11192} \\
    \hline
  \end{tabular}
\end{table*}

\subsection{Ablation Study on Hierarchical Graph Fusion Mechanism}

To verify the effectiveness of HiGFRL's hierarchical graph fusion mechanism, we conducted a comprehensive ablation study. As emphasized in the introduction, existing approaches based on GNNs and reinforcement learning often face the severe challenge of structural-semantic misalignment due to the loose coupling and simple concatenation of topological and resource features. To demonstrate how the hierarchical decoupled design of HiGFRL effectively overcomes this challenge, we designed an ablation baseline named HiGFRL w/o HGF (HiGFRL without Hierarchical Graph Fusion).

HiGFRL w/o HGF completely ablates the hierarchical graph fusion mechanism, degrading the model into a structureless, end-to-end flat GAT. Its core design philosophy is to construct a homogeneous global graph, degrading the hierarchical multi-view design of HiGFRL entirely into a traditional flat structure. The hierarchical graph fusion mechanism is a deeply coupled organic whole, in which the static hypergraph provides global topological priors, the dynamic global graph captures the cluster heterogeneous resource posture, and the local bipartite graph executes precise microscopic matching. To achieve a complete degradation to a generic flat architecture, this baseline model underwent structured ablations and replacements at key stages of information flow. 

Specifically, HiGFRL w/o HGF completely removes the explicit hypergraph convolutional network, abandons the cross-view topological feature fusion strategy, and instead performs graph convolution directly based on raw node attributes on a single homogeneous large graph. This renders the baseline model unable to effectively perceive long-range dependencies such as critical paths and high-order synchronization semantics. Following the cancellation of the hierarchical processing mechanism, the global state encoder used to evaluate the global posture is replaced by a flat graph attention convolution. Due to the lack of cross-view feature decoupling, the topological state of tasks and the underlying physical resource features of VMs are implicitly mixed within a single graph structure. This leads to severe semantic entanglement and over-smoothing of node representations during information passing, which severely limits the evaluation accuracy of the global state evaluator regarding the real physical posture and global load pressure of the complex heterogeneous cluster. Meanwhile, to accommodate the setting of the global homogeneous graph, the original local bipartite graph construction mechanism is removed, and the context-aware matching actor network is downgraded to a global broadcasting and simple concatenation mechanism. This causes the model to lose its ability to focus on fine-grained microscopic matching relationships, failing to explicitly model the deep interactive correlations between tasks and candidate nodes during the feature extraction stage, thereby degrading the matching process into an implicit feature mapping lacking structural dependencies. In summary, this systematic structural degradation ablates the multi-view collaborative representation mechanism, aiming to provide a rigorous ablation baseline to verify the overall effectiveness and necessity of the hierarchical graph fusion mechanism.

\begin{table*}[!t]
  \centering
  \caption{Performance Comparison between HiGFRL and HiGFRL w/o HGF}
  \label{tab:ablation_en}
  \begin{tabular}{p{1.3in}<{\centering} p{1.2in}<{\centering} p{1.2in}<{\centering} p{0.8in}<{\centering} p{0.8in}<{\centering} p{0.7in}<{\centering}}
    \hline
    \textbf{Workload Patterns} & \textbf{Approaches} & \textbf{Makespan} & \textbf{ATFT} & \textbf{ATWT} & \textbf{TRC} \\
    \hline
    \multirow{2}{*}{Standard-50} & HiGFRL w/o HGF & 550714 & 558.39 & 300.53 & \textbf{3} \\
     & HiGFRL & 550714 & \textbf{519.50} & \textbf{276.30} & 13 \\
    \hline
    \multirow{2}{*}{Standard-100} & HiGFRL w/o HGF & 550822 & 772.99 & 432.56 & 454 \\
     & HiGFRL & 550822 & \textbf{722.12} & \textbf{401.86} & \textbf{358} \\
    \hline
    \multirow{2}{*}{Medium-50} & HiGFRL w/o HGF & 552121 & 679.02 & 369.77 & 729 \\
     & HiGFRL & \textbf{551850} & \textbf{597.53} & \textbf{325.39} & \textbf{578} \\
    \hline
    \multirow{2}{*}{Medium-100} & HiGFRL w/o HGF & 636664 & 1861.49 & 1106.39 & 2808 \\
     & HiGFRL & \textbf{561999} & \textbf{1635.77} & \textbf{970.58} & \textbf{2559} \\
    \hline
    \multirow{2}{*}{Large-100} & HiGFRL w/o HGF & 1779568 & 5694.29 & 3412.62 & 8784 \\
     & HiGFRL & \textbf{1198246} & \textbf{5097.41} & \textbf{3069.95} & \textbf{8523} \\
    \hline
    \multirow{2}{*}{Large-250} & HiGFRL w/o HGF & 3950588 & 27088.51 & 17430.96 & 11324 \\
     & HiGFRL & \textbf{3949547} & \textbf{24158.44} & \textbf{15234.66} & \textbf{11192} \\
    \hline
  \end{tabular}
\end{table*}

Table \ref{tab:ablation_en} presents the performance comparison between the complete HiGFRL and the ablated HiGFRL w/o HGF across all six workload patterns. The results demonstrate that the complete hierarchical graph fusion mechanism of HiGFRL provides substantial performance gains, effectively resolving the structural-semantic misalignment problem.

In Standard-50 and Standard-100, both approaches achieve the same Makespan, primarily because the total execution time of small-scale workflows is often limited by the fixed absolute critical path length. However, HiGFRL significantly reduces ATFT and ATWT. In the Standard-100 workload pattern, HiGFRL reduces the ATWT from 432.56 to 401.86. This strongly validates that the local bipartite matching graph introduces a crucial inductive bias, endowing the agent with the ability to make more precise task-to-node pairing decisions, effectively avoiding suboptimal allocations caused by relying on over-smoothed, noisy node features within a flat global graph.

As the workload scale and task graph dependency complexity increase, such as in the Medium-100 and Large-100 scenarios, the performance gap between the two significantly widens. Particularly in the Large-100 scenario, having lost the deep perception of topological structures, HiGFRL w/o HGF suffers severe performance degradation, with its Makespan reaching 1779568. In contrast, HiGFRL maintains a Makespan of 1198246, achieving a substantial reduction of 32.67\%. This proves that the static task dependency hypergraph is indispensable for preserving high-order topological priors. The absence of this module causes severe signal dilution in the flat GAT, rendering the agent unable to identify and prioritize the bottleneck tasks on the critical path hidden deep within the dependency chains.

In medium to large-scale workload scenarios, HiGFRL consistently maintains lower TRC and queueing delays. In the Large-250 scenario, HiGFRL not only effectively reduces delays but also decreases the TRC from 11324 to 11192. This indicates that the hierarchical decoupled design of the dynamic global state representation effectively avoids the mutual interference between task topological features and physical resource features in a flat large graph. Combined with the masked pooling mechanism, HiGFRL can independently and more accurately aggregate the true effective resource load posture across the entire cluster. This effectively prevents the agent from falling into the short-sighted resource over-packing trap under high loads due to distorted state signals, thereby substantially reducing scheduling retries triggered by resource contention.

\section{Conclusion and future work}

In this study, we proposed HiGFRL, a hierarchical graph fusion-driven RL framework, to address instance-level online dependency-aware scheduling in heterogeneous clouds. To overcome the structural-semantic misalignment inherent in existing approaches, HiGFRL explicitly decouples the system state into a static task dependency hypergraph, a dynamic global cluster graph, and a local bipartite matching graph. By integrating these views through a fusion-driven dual-stream architecture and a topology-prior-guided hybrid reward, our approach effectively bridges the gap between logical workflow topology and physical resource posture. Extensive evaluations on real-world Alibaba cluster traces confirm that HiGFRL significantly outperforms baselines, demonstrating exceptional performance in large-scale, high-load scenarios by substantially minimizing Makespan, average task flow time, and average task wait time. 

For future work, we aim to expand our paradigm to multi-cluster federated scheduling, which involves scaling the multi-view state construction to manage complex workload distributions across geographically distributed and privacy-preserving cloud environments.

\bibliographystyle{IEEEtran}
\bibliography{references.bib}

\end{document}